\providecommand{\paperoptions}{amber,normalcite,colorlogo}
\documentclass[\paperoptions]{meta}

\usepackage{amsmath,amssymb}
\usepackage{algorithm}
\usepackage{algorithmic}
\usepackage{url}
\usepackage{xspace}
\usepackage{tabularx,array,colortbl,float}
\usepackage{flafter} 

\title{%
  \texorpdfstring{%
    \raisebox{-0.12em}{\includegraphics[height=1.22em]{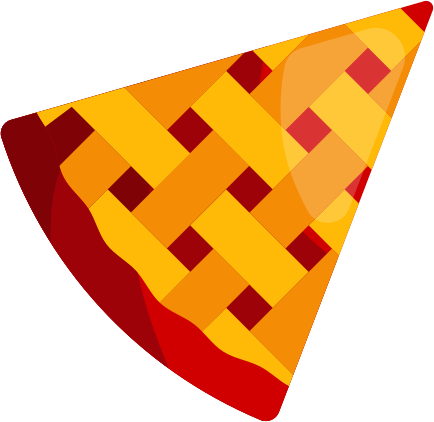}}%
    \hspace{0.18em}%
  }{}%
  MPIE-Bench: Benchmarking Anatomically Plausible Multi-Person Interaction Editing%
}

\author[1,*,\S]{Jiajia Lin}
\author[1,*]{Mingxuan Du}
\author[1]{Tuowen Zhou}
\author[1,2,\ddag]{Benfeng Xu}
\author[1,\dag]{Hongtao Xie}

\affiliation[1]{University of Science and Technology of China}
\affiliation[2]{Metastone Technology, Beijing, China}

\contribution[*]{Equal contribution}
\contribution[\S]{Work done during the internship at Metastone.}
\contribution[\ddag]{Project leader}
\contribution[\dag]{Corresponding author}

\abstract{Text-to-image and personalized editing models now synthesize high-fidelity single-subject images with ease. Yet placing multiple named people into shared contact actions such as embrace, carry, or grapple still exposes major failures: fused limbs, invented extremities, and interpenetrating bodies. Existing evaluations largely overlook these anatomical and geometric issues, and VLM-as-a-judge checklists often saturate on Interaction while the errors remain obvious to humans. We introduce MPIE-Bench, a 2{,}500-sample benchmark of video-mined editing triplets spanning 405 scenes, 14 interaction categories, and four contact densities (C0--C3). We also propose MPIE-Eval, whose two new axes score contact-time geometry from a frozen public multi-person mesh reconstruction. Anatomy asks whether every human-like mass is explained by a complete set of reconstructed bodies, and Interaction asks whether the penetration and surface distance between those bodies match the contact the instruction asked for. Across ten editors, mesh Anatomy tops out at $0.65$ and mesh Interaction at $0.72$ on two different models, so no single editor is strong on both, while VLM checklists rate the same images above $0.95$. A five-rater study confirms that both axes track human judgement more closely than a zero-shot VLM judge, and the rankings hold under ablation of every weight and threshold.
}

\metadata[Code]{\url{https://github.com/AnnLin0628/mpie-bench}}

\begin{document}
\maketitle

\section{Introduction}
\label{sec:intro}

Text-to-image models now compose a coherent scene from a short instruction, follow directions about layout and style, and carry a named person's appearance from a few reference photographs into a new image.
Demand is moving toward scenes where several people interact: two friends embracing, a parent carrying a child, two athletes grappling.
Even the strongest models break down here, and not in the way current benchmarks look for.
The people stay recognizable and so does the requested action; what fails is the body.
Limbs fuse where two people touch, extra hands and feet appear along the contact seam, and one body occupies the volume of another (Figure~\ref{fig:fail}).

Current evaluation cannot see this.
Benchmarks for multi-person and personalized generation ask how many people appear, whether their identities match the references, whether the requested action is present, and how pleasing the result looks~\citep{multihuman2506.20879,withanyone2510.14975,composingpeople2605.23178}, and an image whose bodies are anatomically impossible satisfies all four.
The suites that do ask about interaction often delegate that judgement to a vision-language model~\citep{composingpeople2605.23178}, which in our experiments rates the strongest closed-source models between $0.98$ and $0.99$ on interaction quality for images whose fused limbs are obvious to a human reader.
Whether the requested edit happened and whether the bodies survived it are different questions, and only the first is being asked.
Table~\ref{tab:gap} summarizes the gap.


We close it with \textbf{MPIE-Bench}, a benchmark for multi-person interaction editing, and \textbf{MPIE-Eval}, a protocol that derives contact-time body coherence from 3D human mesh reconstruction rather than from a language model's opinion.
Data comes first, since images of people in dense physical contact are rare in personalization corpora.
We therefore mine video, where frames in which the people stand apart crop into per-person identity references and a later frame in dense contact serves as a held-out target from the same real interaction.
A contact-density curve locates those valleys and peaks, a vision-language model writes the edit instruction backwards from the pair, and human review leaves 2{,}500 triplets over 405 scenes, 14 interaction categories and four contact densities, as the MPIE-Bench section details.

Scoring them is where our main contribution lies.
MPIE-Eval reconstructs the people in a generated image with a frozen, publicly available multi-person mesh recovery model and reads two axes off those meshes.
\textbf{Anatomy} asks whether every human-like mass is explained by a complete set of reconstructed bodies, which is precisely what a fused or invented limb is not.
\textbf{Interaction} asks whether penetration and surface distance between those bodies match the contact the instruction asked for.
The two join the conventional Count, Identity, Instruction and Quality axes and are never blended into a single number; because the frontend and every threshold are fixed in advance, any ranking can be recomputed from released dumps (Figure~\ref{fig:overview}).

Across ten closed- and open-source editors, mesh Anatomy tops out at $0.65$ and mesh Interaction at $0.72$ on two different models, so \emph{no single editor is strong on both}, while a checklist judge rates the same images above $0.95$.
Five-rater human gold confirms that the mesh axes track people more closely than that judge on hard contact items, and the ranking survives ablation of every weight and threshold.

\textbf{Contributions.}
\begin{enumerate}
\setlength{\itemsep}{1pt}
\setlength{\parsep}{0pt}
\setlength{\topsep}{2pt}
\item \textbf{MPIE-Bench}: a licensing-aware mining pipeline and frozen public test set whose references and target come from the same real interaction, stratified by interaction category and contact density.
\item \textbf{MPIE-Eval}: Anatomy and Interaction axes that score contact-time limb completeness and body interpenetration from a frozen public mesh frontend, with every band and weight fixed before scoring.
\item A complete human-consistency study establishing that the protocol is accurate and robust, against five-rater human gold, against a zero-shot VLM judge, and under ablation of every weight, gate and threshold.
\end{enumerate}

\section{Related Work}
\label{sec:related}

\textbf{Multi-person generation and editing.}
Personalized generation began with adapters and spatial conditions binding one identity into a scene.
Group-level editing followed, and the field consolidated around multi-reference in-context editors that take several references and an instruction in a single pass---the interface we evaluate~\citep{multihuman2506.20879,withanyone2510.14975,triopose2606.07053,composingpeople2605.23178,groupdiff2409.14379,inshuman2605.07402,skeleguide2603.01579,psrbench2512.01236,mibe2607.01383}.
Identity preservation and instruction following improve quickly along this line; close-contact geometry does not.

\begin{center}
\small
\setlength{\tabcolsep}{2.2pt}
\begin{tabular}{@{}lccccccc@{}}
\toprule
Benchmark & Count & ID & Anat & Inter & Instr & Qual & VLM \\
\midrule
MultiHuman-Testbench
  & \checkmark & \checkmark & $\times$ & $\triangle$ & \checkmark & \checkmark & \checkmark \\
MIBE
  & \checkmark & \checkmark & $\times$ & $\triangle$ & $\times$ & $\times$ & \checkmark \\
CPT
  & $\times$ & $\times$ & $\times$ & $\triangle$ & \checkmark & $\times$ & \checkmark \\
GroupDiff
  & $\triangle$ & $\times$ & $\times$ & $\times$ & $\triangle$ & $\triangle$ & $\times$ \\
InsHuman
  & $\triangle$ & \checkmark & $\times$ & $\times$ & $\triangle$ & $\triangle$ & \checkmark \\
BodyMetric
  & $\times$ & $\times$ & $\triangle$ & $\times$ & $\times$ & \checkmark & $\times$ \\
\midrule
MPIE-Bench (ours)
  & \checkmark & \checkmark & \checkmark & \checkmark & \checkmark & \checkmark & $\times$ \\
\bottomrule
\end{tabular}
\captionof{table}{Axis coverage of recent multi-person benchmarks ($\checkmark$ scored, $\triangle$ partial, $\times$ absent).
$\triangle$ under Inter means semantic or binding-level interaction only, never contact geometry; $\triangle$ under Anat means single-person anatomy.}
\label{tab:gap}
\end{center}

\textbf{Benchmarks for generation and editing.}
Evaluation followed one step behind: human-aligned protocols for subject-driven generation, large instruction-editing datasets, and the first suites for generating several recognizable humans~\citep{imgedit2505.20275,gedit2603.28547,instructpix2pix2211.09800,dreambenchpp2406.16855,paralleledits2406.00985,micebench2606.icml,cogcanvas2606.15867,multihuman2506.20879}.
Interaction itself entered through stress tests that use physical contact to break identity binding, diagnostics for attribute ownership, and concurrent preference-based interaction binding~\citep{composingpeople2605.23178,mibe2607.01383,cogcanvas2606.15867}.
In every case the interaction question is semantic or relational, never whether the bodies realizing that relation remain anatomically possible.

\textbf{Judging bodies, and judging the judges.}
A parallel line asks whether a generated body is possible at all, from annotated synthetic anatomical defects to learned realism metrics that feed a single-image 3D body representation into a predictor~\citep{abhuman2407.06937,hafbench2605.25759,bodymetric2412.04086,kasbrokenlimbs2026}.
Purpose-built detectors then found that general vision-language models sit near chance at localizing missing or redundant body parts.
All of this work, the reconstruction-assisted metric included, is scoped to a single isolated person.
As Table~\ref{tab:gap} shows, multi-person suites still omit joint anatomical completeness and interaction geometry, and VLM scores on contact axes often stay near ceiling.
MPIE targets this gap with contact-rich editing triplets and mesh-based Anat/Inter (Figure~\ref{fig:overview}).

\textbf{Interaction-aware geometry and HMR.}
Multi-person HMR~\citep{multihmr,condimen2412.13058,intermesh2605.04554}, contact-field representations~\citep{contactfield2024}, hand reconstruction~\citep{hamer2312.06553}, and related contact/distance machinery~\citep{dto2511.13282,gdist2411.11244,sigmagen2510.06469} make geometry-side scoring feasible.
For the public v1 protocol we freeze a single RGB Multi-HMR frontend so editor rankings are reproducible; stronger contact-aware reconstructors can be swapped in later as alternate leaderboard tracks.

\section{MPIE-Bench}
\label{sec:data}

\FloatBarrier
\begin{center}
\includegraphics[width=\linewidth,height=0.38\textheight,keepaspectratio]{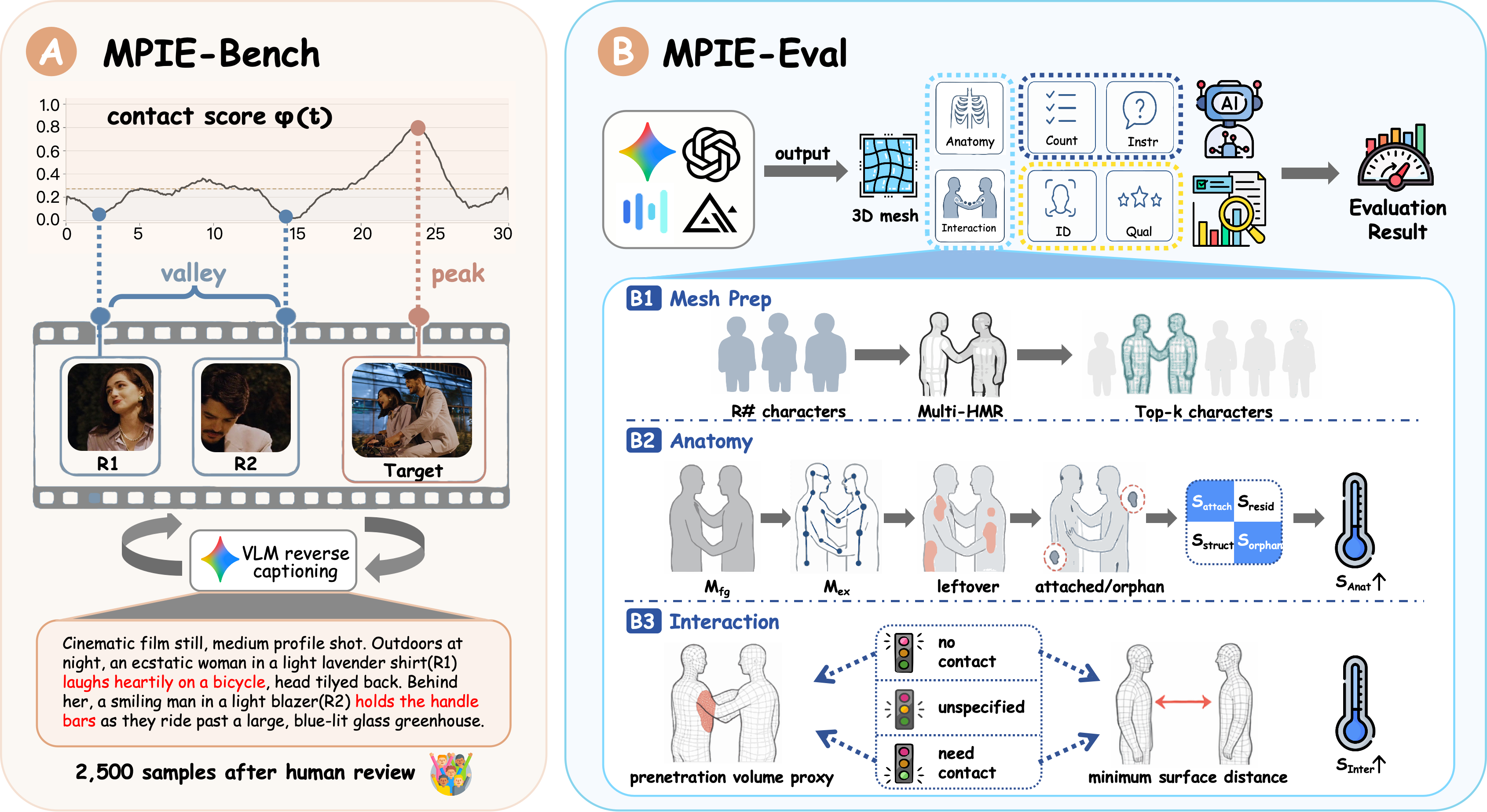}
\captionof{figure}{MPIE-Bench / MPIE-Eval overview. (A)~data construction; (B)~six-axis eval with mesh Anat/Inter.}
\label{fig:overview}
\end{center}

Figure~\ref{fig:overview}(A) sketches construction; Figure~\ref{fig:overview}(B) and MPIE-Eval below define scoring.

\textbf{Sources.} We collect multi-person interaction videos from three open sources: permissively licensed Pexels stock footage, Harmony4D~\citep{harmony4d2410.20294}, and CHI3D~\citep{chi3d2020}. Stock supplies identity and scene diversity; studio sources add hard contact with persistent actor labels. We filter for resolution and multi-person presence, categorize by interaction type, and exclude restricted-tier corpora from the redistributable split. A related CHI3D interaction resource~\citep{flickrci3d2308.01854} is used only to calibrate mesh soft bands on held-out GT reconstructions in MPIE-Eval below; it contributes neither MPIE-Bench test scenes nor any Table~\ref{tab:main} generator output.

\textbf{Input--output selection.} Contact-sparse frames yield clean per-person references; contact-dense frames yield interaction targets. For each video we compute a contact-density curve $\phi(t)$ from person-box IoU, pose proximity, optical flow, and occlusion cues, with refine mix $0.35/0.30/0.20/0.15$ after within-video percentile norm. Local maxima of smoothed $\phi$ with height ${\ge}P_{60}$ and ${\ge}2$\,s gap yield peak targets; a disjoint high-separation pool yields valley references. When multi-person 3D poses are available, peaks/valleys use nearest-joint distance instead. References follow a cross-video $>$ cross-shot $>$ cross-camera $>$ same-shot priority to mitigate copy-paste leakage~\citep{phantomdata2506.18851}.
This valley$\to$peak construction is the operational definition of an editing sample: references are easy-to-crop people, and the target is a harder interaction frame from the same scene narrative, not a random collage.

\textbf{Prompt reverse writing.} Given valley references $\{R_k\}$ and the peak target, a VLM reverse-captions an edit instruction that names roles, the shared action, and contact intent, without revealing target pixels at test time.
Contact intent is later mapped to $\{$required, forbidden, unspecified$\}$ for Interaction mixing (Section~\ref{sec:interaction}); the prompt text itself remains the only instruction shown to editors.

\begin{figure}[!ht]
\centering
\includegraphics[width=0.82\linewidth,height=0.30\textheight,keepaspectratio]{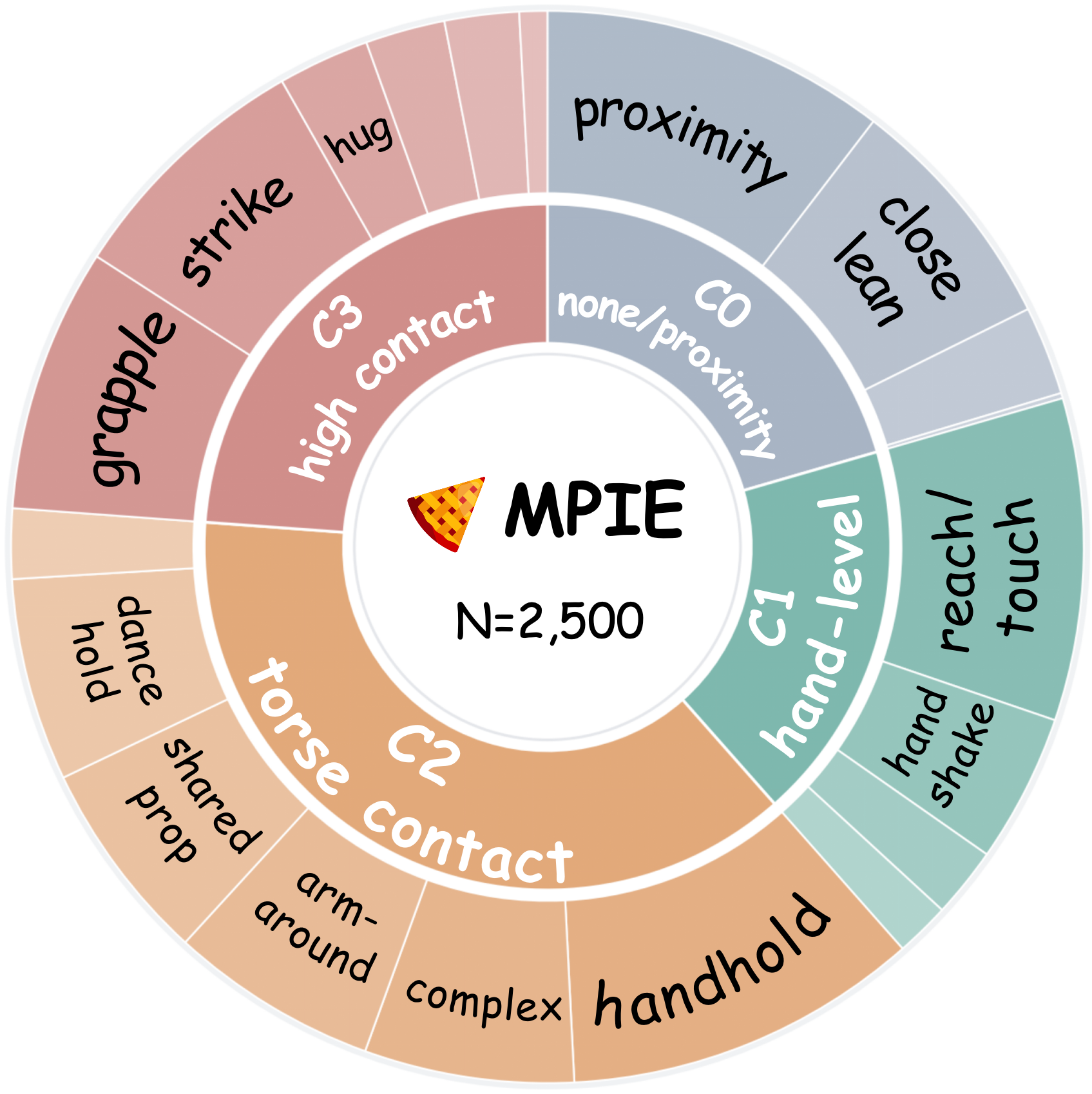}
\caption{Public test-set distribution ($N{=}2{,}500$). Inner ring: contact density from none (C0) to high contact (C3); outer ring: 14 interaction categories.}
\label{fig:distribution}
\end{figure}

\textbf{Human review and distribution.}
Automatic mining yields ${\sim}$20{,}000 candidates.
Adult reviewers remove unsafe or off-task clips, check that each reverse-captioned prompt names the visible interaction, then de-duplicate and cap per scene to freeze 2{,}500 targets across 405 scenes.
Coverage is a $14{\times}4$ grid: fourteen everyday-to-combat interaction categories crossed with four contact-density levels, from no contact through hand-level and torso/point-line contact up to high contact (Figure~\ref{fig:distribution}); about 61\% of samples sit in the two denser bins.
Stock footage supplies most targets (${\approx}$2{,}135); studio CHI3D/Harmony4D add hard-contact coverage, with about 3.2 references per sample on average.
Most prompts expect two people ($2{,}093$ samples); $399$ samples ($16\%$) expect three or more, up to nine.
Splits are scene-level, and reference embeddings freeze at finalize.
Valley$\to$peak may change camera or background within a scene; we score this as reference-conditioned interaction editing under that gap, not as pixel-local inpainting.

\section{MPIE-Eval}
\label{sec:benchmark}

MPIE-Eval reports six axes separately and never blends them into one scalar.
Count, Identity, Instruction, and Quality ask whether the edit request was met; \textbf{Anatomy} and \textbf{Interaction} ask whether bodies stay coherent at contact.
Figure~\ref{fig:overview}(B) sketches the pipeline: reconstruct people once, then score Anatomy from unexplained mass and Interaction from pairwise contact geometry, while the four task axes stay on their own tracks.

\subsection{Why mesh scores for contact geometry}
\label{sec:why-mesh}

VLM checklists often report success on contact while fused limbs and interpenetration remain obvious (Figure~\ref{fig:fail}).
A 2D rubric can confirm that two people are hugging, yet miss whether a limb is invented or one mesh volume occupies another.
We therefore score Anatomy and Interaction from a frozen 3D human reconstruction frontend~\citep{multihmr}: one public stack for all editors, released dumps for audit, and transparent geometric proxies rather than another preference head.
The resulting scores are ranking observables, not physical volumes; soft bands and mix weights below are public and were locked before scoring the ten editors.

\subsection{Reconstruction frontend}
\label{sec:recon}

Every generation $I$ is reconstructed with the same Multi-HMR checkpoint.
We keep the top-$k$ detections by score with $k{=}n_{\mathrm{exp}}$, the expected person count from frozen sample metadata; we do not match identities at this stage.
On $399/2{,}500$ samples ($16\%$) we have $n_{\mathrm{exp}}\ge3$; Interaction still aggregates by the worst kept pair under the same $k$ rule.
If reconstruction fails or returns zero meshes, Anatomy and Interaction are set to $0$ so unscoreable geometry cannot inflate rankings; if fewer people are detected than expected, Interaction is halved.
All Anat/Inter terms below operate on this fixed set of kept meshes.

\subsection{Anatomy}
\label{sec:anat-inter}

Anatomy asks whether every human-like region in the image is explained by a complete set of people.
We group failures into four readable families and score each from $1$ (clean) to $0$ (severe):
\begin{itemize}
\setlength{\itemsep}{1pt}
\setlength{\parsep}{0pt}
\setlength{\topsep}{2pt}
\item \textbf{Extra / fused limbs}: mass glued onto a person that should not be there.
\item \textbf{Floating parts}: detached leftover blobs that belong to no body.
\item \textbf{Body structure}: broken scale, part ownership, or self-collision on the meshes.
\item \textbf{Residual mass}: leftover human-like foreground not covered above.
\end{itemize}

To detect the first two families we compare a classical human-like foreground mask (luma/saturation vs.\ the border; no learned segmenter) with a raster of the kept meshes inside a padded person ROI.
Their mismatch is the leftover fraction
\begin{equation*}
\ell
=
\frac{|M_{\mathrm{fg}}\cap R\setminus M_{\mathrm{ex}}|}{|M_{\mathrm{fg}}\cap R|},
\end{equation*}
and we also count detached leftover blobs $n_{\mathrm{b}}$.
When leftover is high but no blob floats free ($\ell{\ge}0.62$, $n_{\mathrm{b}}{=}0$), we treat it as an attached false limb rather than clothing clutter; otherwise attach terms are down-weighted so loose clothing does not dominate.

Each diagnostic maps through the same soft ramp $s(x;a,b)=\mathrm{clip}((b-x)/(b-a),0,1)$ (good below $a$, fail above $b$).
Within a family we keep the worst sub-score, then mix the four family scores with public weights
\begin{equation*}
S_{\mathrm{anat}}^{\mathrm{mesh}}
=
0.40\,s_{\mathrm{attach}}+0.20\,s_{\mathrm{orphan}}+0.25\,s_{\mathrm{struct}}+0.15\,s_{\mathrm{resid}}.
\end{equation*}
Exact sub-diagnostics and bands (leftover, in-hull enclosure, over-detect, blob count, Multi-HMR scale/ownership/part terms) are frozen in Table~\ref{tab:protocol} and the released scorer; the narrative above is the full scoring logic a reader needs to follow Table~\ref{tab:main}.

\subsection{Interaction}
\label{sec:interaction}

Interaction asks: does reconstructed contact geometry match the prompt's contact intent?
For every kept mesh pair we measure a convex-hull penetration proxy and a surface gap,
\begin{equation*}
\begin{aligned}
V_{p}
&=
r_{\mathrm{in}}\cdot\min\bigl(\mathrm{Vol}_{\mathrm{hull}}(A),\mathrm{Vol}_{\mathrm{hull}}(B)\bigr),\\
d_{s}
&=
\min_{u\in A,\,v\in B}\|u-v\|_{2},
\end{aligned}
\end{equation*}
where $r_{\mathrm{in}}$ is the symmetrized in-hull vertex fraction, used as a ranking proxy rather than an SDF volume.
The same soft map yields penetration and proximity scores; $s_{\mathrm{qual}}$ comes from ownership/over-detect, and forbidden contact uses a clearance score:
\begin{equation*}
\begin{aligned}
s_{\mathrm{pen}}
&=
\min\!\bigl(s(V_p;0.05,0.15),\,s(r_{\mathrm{in}};0.20,0.50)\bigr),\\
s_{\mathrm{prox}}
&=
s(d_s;0.05,0.40),\\
s_{\mathrm{clear}}
&=
1-\max\!\bigl(s_{\mathrm{prox}},\,1{-}s_{\mathrm{pen}},\,1{-}s_{\mathrm{qual}}\bigr).
\end{aligned}
\end{equation*}
We aggregate by the worst pair, then mix by prompt intent mapped to required, forbidden, or unspecified contact:
\begin{equation*}
S_{\mathrm{inter}}^{\mathrm{mesh}} =
\begin{cases}
0.45\,s_{\mathrm{pen}}+0.35\,s_{\mathrm{prox}}+0.20\,s_{\mathrm{qual}} & \text{req.}, \\
0.55\,s_{\mathrm{pen}}+0.45\,s_{\mathrm{clear}} & \text{forb.}, \\
0.85\,s_{\mathrm{pen}}+0.15\,s_{\mathrm{qual}} & \text{unspec.}
\end{cases}
\end{equation*}
Whole-body proximity does not yet enforce contact locus; Section~\ref{sec:consistency} reports a small hand--hand / gaming check that leaves Inter top-2 unchanged.

\subsection{Protocol freeze and calibration}
\label{sec:protocol-freeze}

Table~\ref{tab:protocol} lists the public freeze used for Table~\ref{tab:main}.
Soft Interaction bands are fit on held-out CHI3D-related GT reconstructions~\citep{flickrci3d2308.01854} with no MPIE-Bench test scenes and no generator outputs from the leaderboard: volume/distance cutoffs target GT percentiles so typical contact lies in the soft-ok band and rare extremes approach fail.
We release the recipe and dumps, not restricted calibration pixels.
Two resolution tracks share the same algebra: Track~A at vendor-native resolution and Track~B after letterbox$\to$$1024^{2}$ (Tables~\ref{tab:main}/\ref{tab:resnorm}).

\begin{table}[!t]
\centering
\footnotesize
\setlength{\tabcolsep}{2pt}
\begin{tabular}{@{}p{0.28\linewidth}p{0.64\linewidth}@{}}
\toprule
Quantity & Frozen public value \\
\midrule
Anat mix & $0.40/0.20/0.25/0.15$ (attach/orphan/struct/resid) \\
Attached gate & $\ell{\ge}0.62$, $n_{\mathrm{b}}{=}0$; else attach soft-down $1{-}0.15(1{-}s)$ \\
Anat family mins & attach: $\ell$/fuse/over/extreme; orphan: blob/ofrac; struct: scale/own/part \\
Anat bands & $s(\ell;0.50,0.75)$, $s(r_{\mathrm{in}};0.15,0.45)$; blobs $(0,3)$, ofrac $(0.02,0.15)$ \\
Inter (req.) & $0.45\,s_{\mathrm{pen}}+0.35\,s_{\mathrm{prox}}+0.20\,s_{\mathrm{qual}}$ \\
Inter bands & $V_p\!:\! (0.05,0.15)$; $r_{\mathrm{in}}\!:\! (0.20,0.50)$; $d_s\!:\! (0.05,0.40)$ \\
Tracks & Multi-HMR; A$=$native, B$=$letterbox $1024^{2}$ \\
\bottomrule
\end{tabular}
\caption{Public protocol freeze (pre-scoring). Sub-diagnostics live here; Track~A/B in Tables~\ref{tab:main}/\ref{tab:resnorm}.}
\label{tab:protocol}
\end{table}

\subsection{Task axes}
\label{sec:aux}

We also report four established axes so rankings stay comparable and Anatomy/Interaction cannot be gamed alone.
Two VLM paths are kept separate: a shared geometry checklist (\texttt{vlm\_judge}) and a per-sample Instruction QA bank.

\textbf{Count.} Binary person-count correctness from the shared \texttt{vlm\_judge} checklist, using the same call family as the V-Anat/V-Inter baseline in Experiments, not the Instruction bank.

\textbf{Identity.} Occlusion-aware face matching to frozen reference embeddings with one-to-one assignment and a yaw gate~\citep{multihuman2506.20879,withanyone2510.14975}; GT-visible faces that disappear score zero.

\textbf{Instruction.} A separate pipeline from \texttt{vlm\_judge}: for each of the 2{,}500 samples we freeze a small atomic QA set offline from the edit text alone covering role, asymmetry, and prop. At eval time a VLM only answers on the generation; it cannot invent questions.

\textbf{Quality.} Standard human-preference aesthetic score~\citep{hpsv2}, reported raw and never mixed into Anatomy/Interaction.

\section{Experiments}
\label{sec:exp}


\begin{figure}[!ht]
\centering
\includegraphics[width=\linewidth,height=0.34\textheight,keepaspectratio]{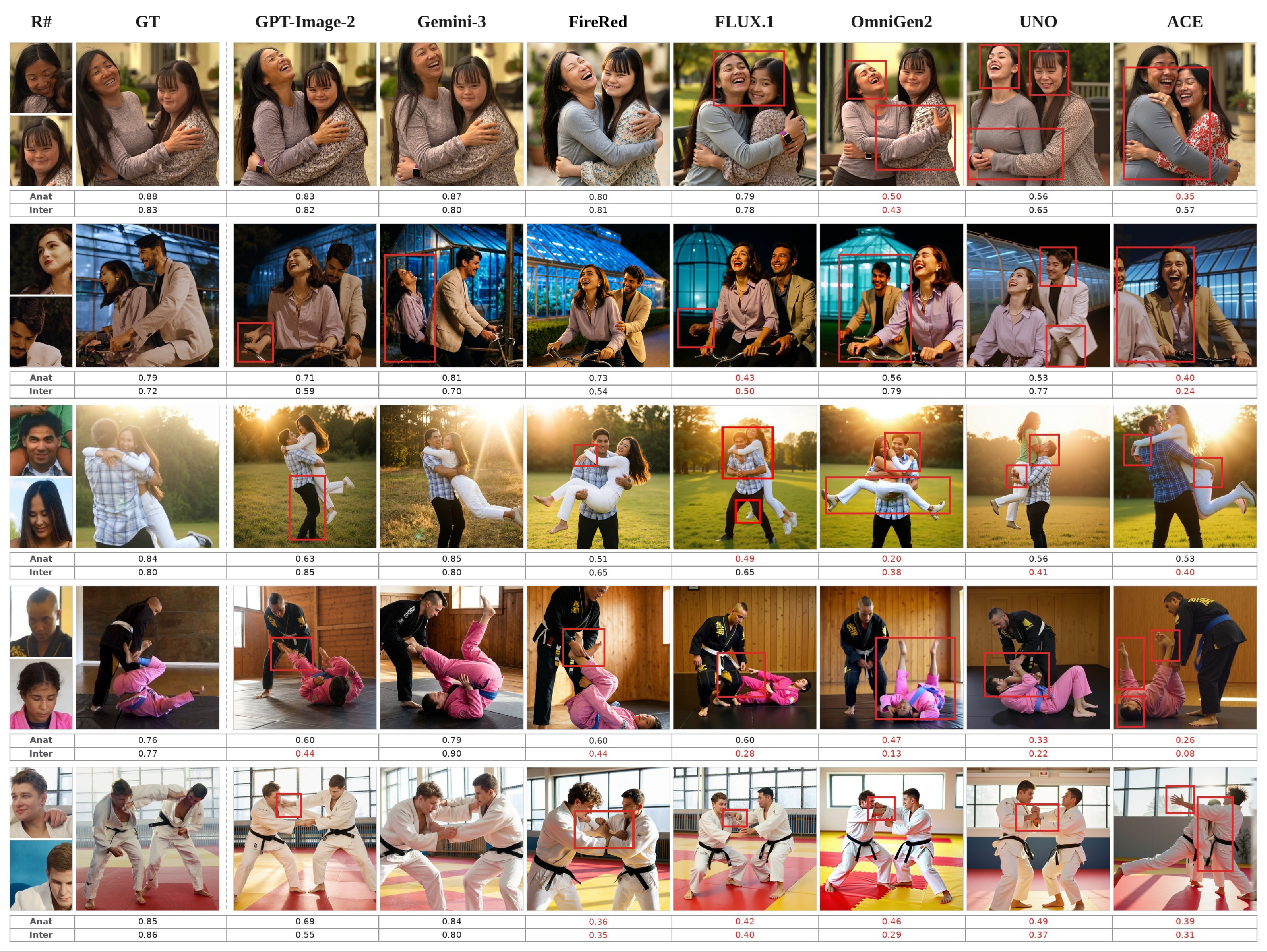}
\caption{Multi-person editing failures. Columns: references, GT, and editor outputs. Red: contact cues in prompts; boxes: limb fusion, missing parts, contact errors.}
\label{fig:fail}
\end{figure}

\begin{table}[!ht]
\centering
\small
\setlength{\tabcolsep}{3.8pt}
\begin{tabular}{@{}lcccccc@{}}
\toprule
Model & \cellcolor{gray!18}Anat & \cellcolor{gray!18}Inter & Count & ID & Instr & Qual \\
\midrule
GPT-Image-2~\cite{gptimage2} & \cellcolor{gray!10}0.58 & \cellcolor{gray!10}0.61 & \textbf{0.74} & \textbf{0.58} & 0.93 & 0.29 \\
Gemini-3-Pro-Image~\cite{gemini3image} & \cellcolor{gray!10}\textbf{0.65} & \cellcolor{gray!10}0.68 & 0.71 & 0.50 & 0.93 & 0.27 \\
Seedream-5-Pro~\cite{seedream5} & \cellcolor{gray!10}0.62 & \cellcolor{gray!10}\textbf{0.72} & 0.72 & 0.49 & \textbf{0.93} & 0.29 \\
\midrule
FLUX.1-Kontext~\cite{fluxkontext} & \cellcolor{gray!10}0.57 & \cellcolor{gray!10}0.57 & 0.69 & 0.21 & 0.71 & 0.30 \\
DreamO~\cite{dreamo} & \cellcolor{gray!10}0.56 & \cellcolor{gray!10}0.57 & 0.69 & 0.20 & 0.72 & 0.30 \\
OmniGen2~\cite{omnigen2} & \cellcolor{gray!10}0.55 & \cellcolor{gray!10}0.53 & 0.69 & 0.17 & 0.77 & 0.29 \\
UNO~\cite{uno} & \cellcolor{gray!10}0.57 & \cellcolor{gray!10}0.53 & 0.68 & 0.05 & 0.65 & 0.28 \\
ACE++~\cite{aceplus} & \cellcolor{gray!10}0.54 & \cellcolor{gray!10}0.45 & 0.49 & 0.02 & 0.54 & 0.29 \\
BAGEL~\cite{bagel} & \cellcolor{gray!10}0.62 & \cellcolor{gray!10}0.59 & 0.68 & 0.15 & 0.75 & 0.26 \\
FireRed-Image-Edit~\cite{firered} & \cellcolor{gray!10}0.63 & \cellcolor{gray!10}0.60 & 0.72 & 0.17 & 0.86 & 0.29 \\
\bottomrule
\end{tabular}
\caption{\textbf{Track~A} (public): full-set six-axis results at vendor-native resolution ($N{=}2{,}500$). Shaded: mesh Anat/Inter. Bold: column best (Qual excluded). Gaps ${\approx}0.01$ read as ties under scene-clustered uncertainty.}
\label{tab:main}
\end{table}

\textbf{Models and settings.}
Ten editors share one multi-reference + instruction interface (7 open, 3 closed: GPT-Image-2~\citep{gptimage2}, Gemini~3~Pro~Image~\citep{gemini3image}, Seedream~5~Pro~\citep{seedream5}; vendor defaults 2026-07; 1 gen/sample, open seed~$0$).
Figure~\ref{fig:fail} and Table~\ref{tab:main} summarize failures and scores.
Two official mesh tracks share Table~\ref{tab:protocol}'s Multi-HMR algebra: Track~A at deployment resolution with vendor-native closed APIs and Table~\ref{tab:settings} for open models; Track~B letterbox$\to$$1024^{2}$ on the full set (Table~\ref{tab:resnorm}).
No model sees the held-out target. Instr uses the frozen QA bank; Count and V-Anat/V-Inter use \texttt{vlm\_judge}.
Unless noted, means are over the full $N{=}2{,}500$ scene-split test set; ${\approx}0.01$ axis gaps read as ties under scene-clustered uncertainty (Section~\ref{sec:consistency}).

\subsection{Main results}
\label{sec:main}

On Table~\ref{tab:main} and Figure~\ref{fig:fail}, closed V-Inter sits at $0.98$--$0.99$ while mesh M-Inter spans $0.45$--$0.72$ (Table~\ref{tab:vlm_mesh}).
Identity shows the steepest closed/open gap ($S_{\mathrm{id}}\approx0.49$--$0.58$ vs.\ $0.02$--$0.21$).
Instruction leadership is closed-dominated, whereas Anat/Inter leaders diverge: Gemini Anat $0.65$, Seedream Inter $0.72$.
Among open editors, FireRed is competitive on Anat but mid-pack on Inter.
For Track~B, Table~\ref{tab:resnorm} re-scores the full set after letterbox$\to$$1024^{2}$; absolute scores rise, and open Anat rank Spearman vs.\ Track~A stays $0.89$ with all-ten $0.83$ and Anat top-3 unchanged. Track~A is the deployment leaderboard; Track~B is the fair-resolution view.

\begin{table}[!t]
\centering
\small
\setlength{\tabcolsep}{2.2pt}
\begin{tabular}{@{}llccc@{}}
\toprule
Model & Backend & Steps & Guid. & Res. \\
\midrule
FLUX.1-Kontext & FLUX.1 & 28 & 2.5 & $1024^2$ \\
DreamO & FLUX.1 & 12 & 4.5 & $1024^2$ \\
OmniGen2 & OmniGen2 & 50 & 5.0/2.0$^\dagger$ & $1024^2$ \\
UNO & FLUX.1$+$LoRA & 25 & 4.0 & $704^2$ \\
ACE++ & FLUX-Fill$+$LoRA & 28 & 50$^\ddagger$ & $768^2$ \\
BAGEL & BAGEL-7B-MoT & 50 & 4.0 & $1024^2$ \\
FireRed-Image-Edit & FireRed-1.1 & 40 & 4.0$^\S$ & $1024^2$ \\
\bottomrule
\end{tabular}
\caption{Open-source inference settings (1 gen/sample; seed~$0$). $^\dagger$text/image guidance; $^\ddagger$ACE \texttt{guide\_scale}; $^\S$FireRed \texttt{true\_cfg\_scale}.}
\label{tab:settings}
\end{table}

\begin{table}[!t]
\centering
\small
\setlength{\tabcolsep}{3pt}
\begin{tabular}{@{}lcccc@{}}
\toprule
Model & V-Anat & V-Inter & M-Anat & M-Inter \\
\midrule
GPT-Image-2 & 0.95 & 0.98 & 0.58 & 0.61 \\
Gemini-3-Pro-Image & 0.95 & 0.99 & 0.65 & 0.68 \\
Seedream-5-Pro & 0.96 & 0.98 & 0.62 & 0.72 \\
\midrule
FLUX.1-Kontext & 0.64 & 0.95 & 0.57 & 0.57 \\
DreamO & 0.71 & 0.96 & 0.56 & 0.57 \\
OmniGen2 & 0.75 & 0.96 & 0.55 & 0.53 \\
UNO & 0.71 & 0.93 & 0.57 & 0.53 \\
ACE++ & 0.56 & 0.88 & 0.54 & 0.45 \\
BAGEL & 0.71 & 0.94 & 0.62 & 0.59 \\
FireRed-Image-Edit & 0.89 & 0.97 & 0.63 & 0.60 \\
\bottomrule
\end{tabular}
\caption{Checklist VLM (V-) vs.\ mesh (M-) Anat/Inter on the full set (Track~A).}
\label{tab:vlm_mesh}
\end{table}

\begin{table}[!t]
\centering
\footnotesize
\setlength{\tabcolsep}{3pt}
\begin{tabular}{@{}lcc@{}}
\toprule
Model & Anat$_B$ & Inter$_B$ \\
\midrule
GPT-Image-2 & 0.681 & 0.754 \\
Gemini-3-Pro-Image & 0.712 & 0.791 \\
Seedream-5-Pro & 0.680 & \textbf{0.803} \\
\midrule
FLUX.1-Kontext & 0.682 & 0.736 \\
DreamO & 0.663 & 0.730 \\
OmniGen2 & 0.678 & 0.750 \\
UNO & 0.691 & 0.732 \\
ACE++ & 0.649 & 0.710 \\
BAGEL & 0.696 & 0.768 \\
FireRed-Image-Edit & \textbf{0.736} & 0.788 \\
\bottomrule
\end{tabular}
\caption{\textbf{Track~B} letterbox$\to$$1024^{2}$ Anat/Inter ($N{=}2{,}500$). Not a drop-in for Table~\ref{tab:main}; vs.\ Track~A Anat $\rho{=}0.89$ (open) / $0.83$ (all).}
\label{tab:resnorm}
\end{table}

\subsection{Contact-density stress}
\label{sec:density}

\begin{figure}[!t]
\centering
\includegraphics[width=\linewidth,height=0.32\textheight,keepaspectratio]{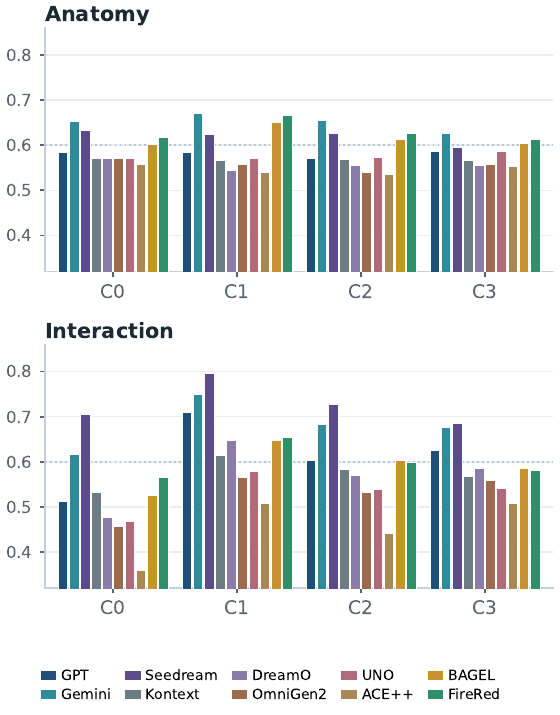}
\caption{Mesh Anat (top) and Inter (bottom) by contact density C0--C3 (full set; $n{=}513/449/942/596$). Point estimates only.}
\label{fig:density}
\end{figure}

Figure~\ref{fig:density} plots Anat/Inter by density (bin sizes $n{=}513/449/942/596$).
Anatomy softens from C2$\to$C3 for Gemini, Seedream, and FireRed, while several open systems stay nearly flat (e.g., DreamO $0.54\to0.56$).
Means often rise from C0 to C1 because C0 mixes non-contact and unspecified-intent samples, so we treat C0 as a control bin.
For high-contact comparisons we prefer absolute Inter under C3 ($\approx0.51$--$0.69$).

\textbf{Protocol stability.}
Offline Anat recompositions (equal weights; $0.30/0.30/0.20/0.20$; attached $\ell\in\{0.50,0.62,0.75\}$) keep ten-model rank Spearman ${\ge}0.98$ vs.\ the public mix; attached gates likewise.
ROI pad $0.22\max(H,W)$ affects only Anat leftover/ROI terms (Inter pad-invariant); a FireRed pad sweep $\{0.15,0.22,0.30\}$ on $N{=}150$ moves Anat by ${+}0.020$/$0$/${-}0.006$.
Track~A recomposes frozen Multi-HMR dumps; Track~B is the official resolution-matched view.
On $N{=}150$, a two-frontend mean (Multi-HMR + alternate HMR) yields Anat/Inter rank Spearman ${\approx}0.92/0.68$ vs.\ Multi-HMR and keeps the same Anat top-3.

\subsection{Metric validation of Anat/Inter}
\label{sec:consistency}

Our metric contribution is Anatomy and Interaction; Count, Identity, Instruction, and Quality follow standard practice. Prior multi-person suites that address contact typically use a VLM or semantic probe (Table~\ref{tab:gap}). We validate mesh Anat/Inter by controlled corruptions and by human agreement against that VLM-style baseline.

\textbf{Sanity checks.} The hull penetration term is a ranking proxy rather than a physical volume. Controlled mesh corruptions on $200$ GT reconstructions lower Interaction under every damage in Table~\ref{tab:corrupt}. A CPU synthetic two-person suite (same Inter algebra; no HMR) passes $6/6$ required direction checks. Keyword contact-intent parsing matches human intent on $145/170$ ($85.3\%$) consistency-pool units; replacing keyword intent with human labels changes full-set Inter means by ${<}10^{-3}$ and leaves the ten-model ranking unchanged.

\begin{table}[!t]
\centering
\small
\setlength{\tabcolsep}{3.5pt}
\begin{tabular}{@{}lccc@{}}
\toprule
Condition & $\Delta$Inter & Direction & $n$ \\
\midrule
Force penetration & $-0.15$ & $\downarrow$ vs.\ baseline & 200 \\
Separate far (required) & $-0.04$ & $\downarrow$ vs.\ baseline & 200 \\
Drop person & $-0.35$ & $\downarrow$ vs.\ baseline & 200 \\
Duplicate person & $-0.14$ & $\downarrow$ vs.\ baseline & 200 \\
\bottomrule
\end{tabular}
\caption{Mesh corruptions on GT reconstructions ($n{=}200$): Inter drops under every damage.}
\label{tab:corrupt}
\end{table}

\textbf{Human agreement vs.\ VLM-style contact scoring.}
On a shared Anat/Inter checklist we compare human gold (H), a frozen mesh$\to$checklist mapping (M), and a VLM judge without fine-tuning on our labels (V). V approximates the contact scoring used by prior multi-person benchmarks; we do not re-evaluate Count/ID/Instruction here.
M fits a small per-item ridge on Multi-HMR fields (e.g., I1 from $s_{\mathrm{pen}}$; Ic from $s_{\mathrm{prox}}$ and $d_s$) and chooses Spearman-max cuts on a held-out label split; cuts freeze before scoring the reported pool and are never applied to Table~\ref{tab:main}'s continuous Anat/Inter.
We use $N{=}170$ hard units (C2--C3); gold is the mean of five adult ratings (mean $\alpha{=}0.53$).
Table~\ref{tab:consist_items} reports per-item Spearman $\rho$: M beats V on 9/10 items (all but person-count I0), with Steiger significance on I3/Ir.
Absolute agreement remains moderate, so we treat the study as a ranking check on hard contact items rather than a per-image oracle.
Editor-level preference anchors in the same pool track mesh Anat/Inter ($\rho\approx0.87/0.80$).

\begin{table}[!t]
\centering
\small
\setlength{\tabcolsep}{2.8pt}
\begin{tabular}{@{}lcccc@{}}
\toprule
Item & $\rho(\mathrm{H},\mathrm{M})$ & CI & $\rho(\mathrm{H},\mathrm{V})$ & CI \\
\midrule
I0 & 0.41 & $[0.28,0.53]$ & \textbf{0.77} & $[0.70,0.82]$ \\
I1 & 0.39 & $[0.25,0.51]$ & 0.30 & $[0.16,0.43]$ \\
Ic & 0.54 & $[0.42,0.64]$ & 0.53 & $[0.41,0.63]$ \\
I3$^\dagger$ & \textbf{0.43} & $[0.30,0.55]$ & 0.09 & $[-0.06,0.24]$ \\
Ir$^\dagger$ & \textbf{0.46} & $[0.33,0.57]$ & 0.17 & $[0.02,0.31]$ \\
A1 & 0.34 & $[0.20,0.47]$ & 0.20 & $[0.05,0.34]$ \\
A2 & 0.24 & $[0.09,0.38]$ & 0.18 & $[0.03,0.32]$ \\
A3 & 0.38 & $[0.24,0.50]$ & 0.21 & $[0.06,0.35]$ \\
A4 & 0.38 & $[0.24,0.50]$ & 0.33 & $[0.19,0.46]$ \\
A5 & 0.26 & $[0.11,0.39]$ & 0.19 & $[0.04,0.33]$ \\
\bottomrule
\end{tabular}
\caption{Per-item checklist Spearman vs.\ human gold ($N{=}170$; Fisher $z$ 95\% CIs). Bold: larger of M/V; $^\dagger$Steiger $p{<}0.05$ for M$>$V. I0 is person-count (VLM-native).}
\label{tab:consist_items}
\end{table}

\textbf{Visible error tags and locus.}
On a hard open-source contact pool ($N{=}220$), geometric failures are dense ($55\%$ have $\ge1$ tag); among anatomy-positive units $63\%$ are \emph{fine-only} (hands $27\%$), so topology checklists under-cover MPIE's regime.
Contact tags move Interaction directionally on that pool.
Public Interaction still uses whole-body proximity: locus-critical prompts (handshake / hand-hold / high-five / arm-around) have pooled Inter ${\approx}0.59$ vs.\ ${\approx}0.56$ elsewhere, while a hand--hand distance swap on a fixed $N{=}150$ subset lowers Inter by only ${\approx}0.005$--$0.011$ overall (${\approx}0.02$--$0.04$ on IDs that enter the hand branch).
A proximity-gaming proxy (high $s_{\mathrm{pen}}$/$s_{\mathrm{prox}}$ but VLM contact-points$=${}no) fires on ${\approx}1.3\%$ of required-ok units and leaves Inter top-2 unchanged; part-conditioned locus remains future work.

\textbf{Uncertainty and sampling.}
Scene-clustered bootstrap gives Seedream$-$Gemini Inter $\Delta{=}{+}0.042$ with CI $[0.019,0.064]$ ($B{=}2000$).
Gaps of ${\approx}0.01$ in Table~\ref{tab:main} read as ties.
Primary tables use one generation per sample; regenerating a fixed $150$-sample subset for three open editors at seeds $\{0,1,2\}$ yields seed-level Anat/Inter stds of the three means in $O(10^{-2})$, well below closed/open and density gaps.
Anat rank Spearman stays $\ge0.98$ under weight/gate ablations.
Per-sample Multi-HMR dumps ship for offline audit; recon failures score Anat/Inter${=}0$ so unscorable geometry cannot inflate the public means.

\section*{Ethical Statement}
\label{sec:ethics}

\textbf{Data licensing.} Harmony4D pixels ship under MIT. Public Pexels stock follows the Pexels License for IDs, URLs/timestamps, scripts, and derived crops for research, with no re-host as a stock library. Academically licensed sources such as CHI3D ship as IDs/timestamps/boxes/scripts under original terms; the CHI3D interaction resource~\citep{flickrci3d2308.01854} calibrates thresholds only, with the recipe released and not the pixels. Restricted-tier corpora never enter the public suite.

\textbf{Safety and intended use.} Keyframes pass automated minor detection and flagged frames are dropped; non-sexualization constraints apply to identity-conditioned generation; combat/restrain prompts are sports- or self-defense-style. Checklist labels used adult raters only. Sources are appearance-only public/academic likenesses; we scope out placing identifiable real individuals into synthetic scenes without consent. Metrics, including ID, are for research diagnosis, not rewards for non-consensual likenesses; downstream users must obey source licenses and any local consent rules.

\section{Conclusion}
\label{sec:conclusion}

We set out to make contact-time body coherence measurable for multi-person interaction editing.
MPIE-Bench contributes a licensing-aware, scene-split test set of 2{,}500 video-mined triplets over 405 scenes, stratified by 14 interaction categories and C0--C3 contact density, with held-out targets so editors cannot copy pixels.
MPIE-Eval contributes a six-axis protocol that keeps Count/ID/Instruction/Quality for task success while scoring Anatomy/Interaction from a frozen, dump-auditable Multi-HMR frontend under dual public tracks: Track~A at vendor-native resolution and Track~B after letterbox$\to$$1024^{2}$.

Across ten editors, the mesh axes do work the checklists do not: closed V-Inter saturates near ceiling while M-Inter retains dynamic range; identity remains the steepest closed/open gap; Anat and Inter leaders need not coincide; and density, corruption, human-checklist, and ablation checks support model-level ranking under the frozen frontend rather than a single blended score.

We release the redistributable subset, evaluation code, per-sample Multi-HMR dumps, and the calibration recipe so the community can audit rankings and add alternate-frontend tracks without redefining the editing task.

\section*{Acknowledgments}
We thank collaborators who helped with data review and annotation tooling.
This preprint presents the frozen evaluation protocol, full-set tables, and
appendix analyses in a single-column format without a page cap.

\bibliographystyle{assets/plainnat}
\bibliography{references}

\beginappendix

\section{Extended protocol notes}
\label{app:protocol}

\paragraph{Instr~v2 design.}
We separate question generation from answering. Offline, a text-only stage extracts atomic claims from the edit instruction and freezes a per-sample QA bank with buckets \texttt{role}/\texttt{asymm}/\texttt{prop} (main) and optional \texttt{scene} (diagnostic only), forbidding interaction-occurrence, count, face-identity, and anatomy questions. At evaluation time the VLM answers the frozen questions from the generated image alone and cannot rewrite them. An earlier image-conditioned self-authored \texttt{instr\_qa} protocol saturated on closed-source models and is deprecated.

\paragraph{Mesh Interaction intent.}
Contact intent is read from the edit prompt text (not the dataset category label), yielding $\mathrm{required}$ / $\mathrm{forbidden}$ / $\mathrm{unspecified}$. Penalties combine interpenetration volume, enclosure fraction, proximity, and contact-quality terms , with soft ramps calibrated on GT frames. The metric flags unreasonable geometry; higher Inter than GT can occur when a generation under-penetrates relative to real contact.

\paragraph{Release artifacts.}
We release: (i)~the redistributable test subset and scene-level split; (ii)~frozen reference face embeddings; (iii)~evaluation scripts for the six-axis protocol including 3D mesh Anatomy/Interaction; (iv)~the human checklist template and agreement scripts; (v)~the metric calibration recipe for GT percentile thresholds.

\section{Toward anatomy-aware training (future work)}
\label{app:training}

The main paper is evaluation-only. For completeness we sketch a training recipe we are exploring on top of MPIE-Bench signals; none of these results are required to use the benchmark.

\paragraph{Base model.}
Fine-tune FLUX.1~Kontext~\citep{fluxkontext} with a rank-512 LoRA rather than full-parameter updates, because (a)~narrow-task data at $\sim10^4$ scale risks catastrophic forgetting under full fine-tuning of a large DiT, and (b)~Kontext is guidance-distilled; full updates can disturb the distillation structure. Kontext already exposes a native reference-conditioning pathway, so we do not add an identity encoder unless post-hoc ID/$M_{CP}$ diagnostics indicate copy-paste cheating.

\paragraph{Auxiliary skeleton stream.}
A lightweight auxiliary LoRA + projection head predicts a 2D multi-person skeleton visualization during training (discarded at inference), encouraging the shared backbone to keep one complete skeleton per person under occlusion -- targeting limb-fusion and limb-attribution failures.

\paragraph{Contact-density curriculum.}
Training samples are scheduled along C0--C3: early C0--C1, mid C2, late C3, so the model stabilizes body completeness before deep mutual occlusion. This is a dataloader-level change with zero architectural cost.

\paragraph{What we defer.}
Corrupted positive/negative pairs from metric calibration are reserved for future preference optimization / RL stages (e.g., Diffusion-DPO~\citep{diffusiondpo2311.12908}, Flow-GRPO~\citep{flowgrpo2505.05470}) rather than standard supervised losses that would simply reproduce corruptions.

\section{VLM-as-judge saturation diagnostic}
\label{app:vlm-sat}

An all-checklist visual judge yields Anatomy/Interaction means near $0.98$--$0.99$ for strong closed-source editors while fused limbs and pathological interpenetration remain visible. Parametric joint/bone priors similarly saturate ($\approx0.99$) because hallucinated extremities often lie outside the fitted topology. Primary tables therefore report mesh-anchored Anatomy and prompt-conditioned Interaction in 3D mesh space; checklist Anat/Inter remain diagnostics only and are released with the evaluation code rather than used for ranking.

\section{Coverage rules}
\label{app:coverage}

An axis is reported when it meets the coverage rule in the release scripts (minimum fraction of scored generations per model and axis). All eight systems in the main tables meet that rule for every reported axis.

\section{End-to-end pipeline (Algorithm 1)}
\label{supp:alg}

This supplement freezes the numeric constants, sensitivity checks, and human-consistency details used by the mining and mesh Anat/Inter pipeline described in the main paper.

\begin{algorithm}[tb]
\caption{MPIE-Bench pipeline (data construction + Anat/Inter scoring)}
\label{alg:pipeline}
\small
\textbf{Input}: video corpus $\mathcal{V}$; editor $f$; frozen pack $\mathcal{D}$\\
\textbf{Output}: per-sample six-axis scores (Anat/Inter detailed below)
\begin{algorithmic}[1]
\STATE \textbf{// A. Mine reference--instruction--target triplets}
\FOR{video $v\in\mathcal{V}$}
\STATE Coarse $\phi_{\mathrm{coarse}}$ then refine $\phi$ (IoU/Prox/Flow/Occ); peaks via smoothed $\phi$ ($P_{60}$, $\ge2$\,s)
\STATE $\mathcal{T}\leftarrow\mathrm{TopPeaks}(\phi)$; $\mathcal{R}\leftarrow$ high-separation / low-overlap pool (disjoint from $\mathcal{T}$)
\STATE Pick peak target crop; valley identity crops (tiered anti-copy-paste); emit prompt $p$, $n_{\mathrm{exp}}$, C0--C3
\ENDFOR
\STATE Human-review scenes; freeze public split $\mathcal{D}$ ($N{=}2{,}500$) and reference embeddings
\STATE \textbf{// B. Generate}
\FOR{sample $(I_{\mathrm{refs}},p)\in\mathcal{D}$}
\STATE $I\leftarrow f(I_{\mathrm{refs}},p)$ \COMMENT{identical refs/prompt across editors}
\STATE Score Count / ID / Instr / Qual with checklist, ArcFace, HPSv2
\STATE \textbf{// C. Mesh Anat / Inter}
\STATE Reconstruct meshes on $I$ (frozen Multi-HMR); keep top-$n_{\mathrm{exp}}$ detections
\STATE Build leftover / enclosure / ownership proxies
\STATE Form $s_{\mathrm{attach}},s_{\mathrm{orphan}},s_{\mathrm{struct}},s_{\mathrm{resid}}$ ($s{=}1{-}P$)
\STATE $S_{\mathrm{anat}}\leftarrow\sum w\,s$ with weights $0.40/0.20/0.25/0.15$
\STATE Measure $V_{p},r_{\mathrm{in}},d_{s}$ $\to$ $s_{\mathrm{pen}},s_{\mathrm{prox}},s_{\mathrm{qual}}$
\STATE $\mathrm{intent}\leftarrow\mathrm{ParseContactIntent}(p)$
\STATE \textbf{if} required \textbf{then} $S_{\mathrm{inter}}\leftarrow0.45\,s_{\mathrm{pen}}+0.35\,s_{\mathrm{prox}}+0.20\,s_{\mathrm{qual}}$
\STATE \textbf{else if} forbidden \textbf{then} $S_{\mathrm{inter}}\leftarrow0.55\,s_{\mathrm{pen}}+0.45\,s_{\mathrm{clear}}$
\STATE \textbf{else} $S_{\mathrm{inter}}\leftarrow0.85\,s_{\mathrm{pen}}+0.15\,s_{\mathrm{qual}}$
\STATE \textbf{if} $n_{\mathrm{humans}}<n_{\mathrm{exp}}$ \textbf{then} $S_{\mathrm{inter}}\leftarrow\tfrac12 S_{\mathrm{inter}}$
\ENDFOR
\end{algorithmic}
\end{algorithm}

\section{Dataset distribution}
\label{supp:dist}

The main paper includes the public test-set sunburst (density$\times$category). Source-stratified counts and per-class breakdowns ship with the release tables.

\section{Frozen metric constants}
\label{supp:cal}

Public rankings freeze Multi-HMR backend \texttt{multiHMR\_896\_L}. Table~\ref{tab:supp_const} lists the constants used by the public Anat/Inter scorers and the mining defaults stated in the main paper.

\begin{table}[ht]
\centering
\small
\setlength{\tabcolsep}{3.5pt}
\begin{tabular}{@{}llp{0.58\textwidth}@{}}
\toprule
Group & Symbol / name & Value / definition \\
\midrule
Mining (coarse) & $\phi_{\mathrm{coarse}}$ & $0.7\,\mathrm{IoU}_{\mathrm{pair}}+0.3\,E_{\mathrm{mot}}$ \\
Mining (refine) & $(w_{\mathrm{IoU}},w_{\mathrm{Prox}},w_{\mathrm{Flow}},w_{\mathrm{Occ}})$ & $(0.35,0.30,0.20,0.15)$ after within-video percentile norm \\
Peak detect & smooth / height / gap & $3$-frame box; $\ge P_{60}(\phi)$; $\ge2$\,s separation \\
Valley refs & --- & high-separation / low-overlap pool; disjoint from peaks \\
Face gate & det / yaw & $\ge0.62$; $|\mathrm{yaw}|\le38^{\circ}$ \\
\midrule
Foreground $M_{\mathrm{fg}}$ & classical mask & $|L-b|>0.10$ or $(S>S_{70}\land$ person-side of mid-luma); no learned segmenter \\
Explained / ROI & $M_{\mathrm{ex}},R$ & thick skeleton/torso of kept bodies; joint-box pad $0.22\max(H,W)$, floor $24$\,px \\
Leftover / blobs & $\ell,n_{b}$ & $\ell=|M_{\mathrm{fg}}\!\cap\!R\setminus M_{\mathrm{ex}}|/|M_{\mathrm{fg}}\!\cap\!R|$; $n_{b}$ detached leftover blobs (frac$\ge0.025$) \\
$s_{\ell}$ / $s_{\mathrm{fuse}}$ & leftover / in-hull & $s(\ell;0.50,0.75)$; $s(r_{\mathrm{in}};0.15,0.45)$ \\
Attached signature & gate on $s_{\ell},s_{\mathrm{fuse}}$ & if $\ell\ge0.62$ and $n_{b}=0$: use as-is; else $s\leftarrow1-0.15(1-s)$ \\
$s_{\mathrm{over}}^{\mathrm{gated}}$ / $s_{\mathrm{extreme}}$ & over-detect & gated soft band on $n_{\mathrm{raw}}/n_{\mathrm{exp}}$ when attach evidence; milder ungated floor if $n_{\mathrm{raw}}{\gg}n_{\mathrm{exp}}$ \\
$s_{\mathrm{blob}}$ / $s_{\mathrm{ofrac}}$ & orphan & blob count $(0,3)$; orphan-frac $(0.02,0.15)$ \\
$s_{\mathrm{scale}},s_{\mathrm{own}}^{\mathrm{amp}},s_{\mathrm{part}}$ & structure & Multi-HMR scale / ownership amp~$2.0$ / part self-collision \\
$s_{\mathrm{resid}}$ & residual & foreground explain-residual \\
Anat mix & $S_{\mathrm{anat}}$ & $0.40/0.20/0.25/0.15$ on $s_{\mathrm{attach}},s_{\mathrm{orphan}},s_{\mathrm{struct}},s_{\mathrm{resid}}$ \\
\midrule
$s_{\mathrm{pen}}$ & $V_p$, pair $r_{\mathrm{in}}$ & $\min\!\bigl(s(V_p;0.05,0.15),\,s(r_{\mathrm{in}};0.20,0.50)\bigr)$ \\
$s_{\mathrm{prox}}$ & gap $d_s$ & $s(d_s;0.05,0.40)$ \\
$s_{\mathrm{qual}}$ / $s_{\mathrm{clear}}$ & quality / clearance & ownership$+$over-detect; $s_{\mathrm{clear}}{=}1{-}\max(s_{\mathrm{prox}},1{-}s_{\mathrm{pen}},1{-}s_{\mathrm{qual}})$ \\
Inter mix & intent cases & req.\ $0.45/0.35/0.20$; forb.\ $0.55/0.45$; unspec.\ $0.85/0.15$ \\
Missing person & Inter & $\times\tfrac12$ if $n_{\mathrm{humans}}<n_{\mathrm{exp}}$ \\
$V_p$ & hull proxy & $r_{\mathrm{in}}\cdot\min(\mathrm{Vol}_{\mathrm{hull}}A,\mathrm{Vol}_{\mathrm{hull}}B)$; not SDF \\
\bottomrule
\end{tabular}
\caption{Frozen constants for mining gates and mesh Anat/Inter (complete soft-score sheet).}
\label{tab:supp_const}
\end{table}

\section{Anatomy and Interaction formulas}
\label{supp:anat-inter}

The main paper states leftover $\ell$, the Anatomy soft-score structure, the Inter pair proxies $V_p$/$d_s$, and the public weighted mixes; here we expand operational sub-terms, aggregation, and the hull-proxy rationale. Reconstruct with frozen Multi-HMR; keep top-$k$ with $k=n_{\mathrm{exp}}$. On generation $I$, let $M_{\mathrm{fg}}$ be a classical human-like foreground, $M_{\mathrm{ex}}$ thick skeleton/torso coverage of kept bodies, and $R$ the joint-box union padded by $0.22\max(H,W)$ (floor $24$\,px). Leftover fraction and detached-blob count are
\begin{equation*}
\ell=\frac{|M_{\mathrm{fg}}\cap R\setminus M_{\mathrm{ex}}|}{|M_{\mathrm{fg}}\cap R|},\qquad
n_{b}=\#\{\text{detached leftover blobs, frac}\ge0.025\}.
\end{equation*}
Additional constants appear in Table~\ref{tab:supp_const}. Anatomy soft scores (same algebra as the main paper; $s{=}1{-}P$ is only an implementation alias):
\begin{align*}
s_{\mathrm{attach}}
&=
\min\!\bigl(s_{\ell},\,s_{\mathrm{fuse}},\\
&\qquad s_{\mathrm{over}}^{\mathrm{gated}},\,s_{\mathrm{extreme}}\bigr),\\
s_{\mathrm{orphan}}
&=
\min\!\bigl(s_{\mathrm{blob}},\,s_{\mathrm{ofrac}}\bigr),\\
s_{\mathrm{struct}}
&=
\min\!\bigl(s_{\mathrm{scale}},\,s_{\mathrm{own}}^{\mathrm{amp}},\,s_{\mathrm{part}}\bigr),
\end{align*}
where $s_{\ell}$/$s_{\mathrm{fuse}}$ band $\ell$ and $r_{\mathrm{in}}$; $s_{\mathrm{over}}^{\mathrm{gated}}$ bands $n_{\mathrm{raw}}/n_{\mathrm{exp}}$ when attach evidence is present; $s_{\mathrm{extreme}}$ is an ungated mild floor for extreme over-detect; $s_{\mathrm{blob}}$/$s_{\mathrm{ofrac}}$ score detached leftovers; $s_{\mathrm{scale}}$, $s_{\mathrm{own}}^{\mathrm{amp}}{=}1{-}\min(1,(1{-}s_{\mathrm{own}}){\cdot}2)$, and $s_{\mathrm{part}}$ come from Multi-HMR scale / ownership / part-mesh scores; $s_{\mathrm{resid}}$ is foreground explain-residual.
Leftover/fuse enter the $\min$ as-is only under the attached signature ($\ell\ge0.62$, $n_{b}=0$); otherwise replace $s$ by $1{-}0.15(1{-}s)$ (equivalent to scaling the old penalty by $0.15$). Then
\begin{equation*}
\begin{aligned}
S_{\mathrm{anat}}^{\mathrm{mesh}}
&=
0.40\,s_{\mathrm{attach}}+0.20\,s_{\mathrm{orphan}}\\
&\quad+0.25\,s_{\mathrm{struct}}+0.15\,s_{\mathrm{resid}}.
\end{aligned}
\end{equation*}
For Interaction, each kept mesh pair yields
\begin{align*}
V_{p}
&=
r_{\mathrm{in}}\cdot\min\bigl(\mathrm{Vol}_{\mathrm{hull}}(A),\mathrm{Vol}_{\mathrm{hull}}(B)\bigr),\\
d_{s}
&=
\min_{u\in A,\,v\in B}\|u-v\|_{2},
\end{align*}
$s_{\mathrm{pen}}$ from the stricter of volume and $r_{\mathrm{in}}$ bands, $s_{\mathrm{prox}}$ from $d_{s}$, $s_{\mathrm{qual}}$ from ownership/over-detect, and
\begin{equation*}
S_{\mathrm{inter}}^{\mathrm{mesh}} =
\begin{cases}
0.45\,s_{\mathrm{pen}}+0.35\,s_{\mathrm{prox}}+0.20\,s_{\mathrm{qual}} & \text{req.}, \\
0.55\,s_{\mathrm{pen}}+0.45\,s_{\mathrm{clear}} & \text{forb.}, \\
0.85\,s_{\mathrm{pen}}+0.15\,s_{\mathrm{qual}} & \text{unspec.}
\end{cases}
\end{equation*}
halved if $n_{\mathrm{humans}}<n_{\mathrm{exp}}$.
Across $\ge2$ kept bodies, pair diagnostics aggregate by worst contact geometry before the mix above: $\max$ over pairs of penetration/enclosure proxies and $\min$ over pairs of surface distance (one severely intersecting pair dominates the unit score).

\paragraph{Exception handling.}
Expected count $n_{\mathrm{exp}}$ comes from the frozen sample metadata. Multi-HMR detections are sorted by detector score; we keep the top-$k$ with $k{=}n_{\mathrm{exp}}$ (no identity-aware matching in the public protocol). If reconstruction fails or yields zero usable meshes, Anat and Inter for that unit are $0$. If $0<n_{\mathrm{det}}<n_{\mathrm{exp}}$, Interaction is halved (missing-person penalty) and Anatomy applies the under-detection path in the Anat composer. Surplus detections beyond $k$ are ignored for pairing. Intent for the Interaction mix is keyword-parsed from the edit instruction (human overrides on the consistency pool do not change full-set rankings; main text).

\paragraph{Why hull-normalized $V_p$.}
Exact mesh--mesh penetration via SDF or signed volumes is accurate but slow and brittle under Multi-HMR topology noise. We use
\begin{equation*}
V_{p}=r_{\mathrm{in}}\cdot\min\bigl(\mathrm{Vol}_{\mathrm{hull}}(A),\mathrm{Vol}_{\mathrm{hull}}(B)\bigr):
\end{equation*}
$r_{\mathrm{in}}$ captures relative enclosure while the hull factor removes absolute mesh units. Outstretched limbs inflate convex hulls and can overstate fusion; we therefore treat $V_p$ as a \emph{ranking} observable under a frozen reconstructor, not a metric contact volume, and calibrate soft cutoffs on held-out GT frames in the same units. The CPU synthetic suite in Section~\ref{supp:corruption} includes a hull-inflation stress case (outstretched arms) that lowers the effective fusion signal relative to a part-aware volume oracle, confirming the known bias direction without changing editor rankings under the frozen frontend.

\section{Mesh$\to$checklist mapping (M)}
\label{supp:mesh_map}

For each checklist item $q$, hand-select a small feature vector $\mathbf{x}_q$ of related Multi-HMR fields (e.g., I1: $s_{\mathrm{pen}}$ and penetration ratios; A1: leftover/$P_{\mathrm{extra}}$/$s_{\mathrm{own}}$; Ic: $s_{\mathrm{prox}},d_s$). On a held-out calibration split of human labels $y_q$, $z$-score features to $\tilde{\mathbf{x}}_q$ and fit a ridge regressor
\begin{align*}
\hat{y}_q
&=
\mathbf{w}_q^{\!\top}\tilde{\mathbf{x}}_q+b_q,\\
(\mathbf{w}_q,b_q)
&=
\arg\min_{\mathbf{w},b}\,
\sum_{i}
\bigl(y_q^{(i)}-\mathbf{w}^{\!\top}\tilde{\mathbf{x}}_q^{(i)}-b\bigr)^{2}\\
&\qquad+\lambda\|\mathbf{w}\|_{2}^{2}.
\end{align*}
Discretize $\hat{y}_q$ with cuts that maximize Spearman correlation with $y_q$ on the calibration split. Binary items use a (possibly polarity-inverted) threshold $\tau_q$; ternary items use ordered cuts $\tau_q^{(1)}<\tau_q^{(2)}$:
\begin{equation*}
c_q(\hat{y})
=
\begin{cases}
\mathbf{1}[\hat{y}\ge\tau_q] & \text{binary},\\
0 & \hat{y}<\tau_q^{(1)},\\
1 & \tau_q^{(1)}\le\hat{y}<\tau_q^{(2)},\\
2 & \text{else (ternary)}.
\end{cases}
\end{equation*}
Weights, biases, and cuts freeze before evaluating the $N{=}170$ consistency pool; Interaction items apply the same intent-conditioned dependency rules as humans.

\section{Two VLM paths: checklist vs.\ Instruction QA}
\label{supp:vlm_paths}

MPIE-Eval uses two \emph{disjoint} VLM protocols. Conflating them is a common reading error, so we state the split once:

\begin{center}
\small
\setlength{\tabcolsep}{4pt}
\begin{tabular}{@{}p{0.22\textwidth}p{0.35\textwidth}p{0.35\textwidth}@{}}
\toprule
& Shared checklist (\texttt{vlm\_judge}) & Per-sample Instruction QA \\
\midrule
Main-table use & Count; also V-Anat / V-Inter as a saturation baseline vs.\ mesh & Instr only \\
Questionnaire & One fixed field set for all samples (count, anatomy tags, semantic / contact / penetration, \ldots) & One frozen atomic bank per sample (2{,}500 banks) \\
Authoring & Protocol fields are written once; judge fills them on the generation (+ refs) & Questions authored from \emph{text only}; answering sees the generation \\
Not used for & Main-table Instr & Count / V-Anat / V-Inter \\
\bottomrule
\end{tabular}
\end{center}

Sections~\ref{supp:vlm}--\ref{supp:instr} detail each path. Main-table Instr never reads checklist Instruction-like fields from \texttt{vlm\_judge}.

\section{Checklist VLM protocol (\texttt{vlm\_judge})}
\label{supp:vlm}

The main-paper Count column and the V-Anat / V-Inter baseline columns use a frozen shared checklist \texttt{vlm\_judge} with a frontier vision-language judge.
This is \emph{not} the main-table Instruction score (Section~\ref{supp:instr}).
Per sample the judge sees the generation plus ordered reference crops $R_1,\ldots,R_n$ (no GT target at test time) and structured metadata (instruction, expected person count, interaction type, contact density). Outputs are binary/ordinal checklist items, not Likert scores:
\begin{itemize}
\setlength{\itemsep}{1pt}
\setlength{\parsep}{0pt}
\item \textbf{Count:} whether the visible person count matches $n_{\mathrm{exp}}$ (feeds the main-table Count axis).
\item \textbf{Anatomy (V-Anat):} closed-set error tags (e.g., extra/missing limb, merged body, limb ownership, bad hand, floating part); score $1$ if empty, else $1-\min(1,0.25\,n_{\mathrm{err}})$.
\item \textbf{Interaction (V-Inter):} three cuts---semantic action presence, contact-point fit, and no pathological penetration---each $\mathrm{yes}{=}1$/$\mathrm{partial}{=}0.5$/$\mathrm{no}{=}0$, then averaged. Real deep contact is not scored as penetration.
\end{itemize}

\paragraph{Visible contract (fields + example).}
Public Count / V-Anat / V-Inter read only the fixed fields below (same schema for every sample). Closed Anat vocabulary:
\texttt{extra\_limb}, \texttt{missing\_limb}, \texttt{merged\_body}, \texttt{limb\_ownership\_error}, \texttt{bad\_hand}, \texttt{impossible\_joint}, \texttt{floating\_part}, \texttt{face\_melt}.
A real judge return used for the V-Anat/V-Inter baseline (excerpt; identity fields omitted here because the public ID axis uses ArcFace, not this checklist):
\begin{verbatim}
{
  "count_detected": 2,
  "count_expected": 2,
  "count_pass": true,
  "anat_errors": [
    {"type": "bad_hand",
     "where": "high-five contact region fingers appear fused/overcrowded"}
  ],
  "anat_pass": false,
  "inter": {
    "semantic": "yes",
    "contact_points": "yes",
    "no_pathological_penetration": "partial"
  }
}
\end{verbatim}
Prompt skeleton (abridged): Image~1 $=$ generation; Images~$2{\ldots}$ $=$ ordered references; text carries instruction, $n_{\mathrm{person}}$, interaction type, contact density, and contact-point hints; ``return only one JSON object''; deep legitimate contact is not pathological penetration. Full prompt text and schema checks ship with the release scripts.
Any Instruction-like free-form fields that an older joint call might have emitted are \emph{ignored} for the public leaderboard (main-table Instr uses Section~\ref{supp:instr}).
Closed-source V-Inter stays near ceiling ($0.98$--$0.99$) while M-Inter spans $0.45$--$0.72$; closed-source V-Anat is high but not saturated ($0.95$--$0.96$) against M-Anat $0.54$--$0.65$.

\section{Instruction QA protocol}
\label{supp:instr}

Main-table Instruction is a two-stage text-then-answer bank (TIFA-style), separate from \texttt{vlm\_judge}.

\textbf{Stage A (offline; text only).}
For each of the 2{,}500 edit prompts we freeze a small atomic question set with gold answers implied by the text (role binding to reference slots, asymmetric duties, prompt-mandated props/attributes). Question authoring never sees the generation or reference pixels, so the bank cannot drift with the editor under test. Contact-geometry, person-count, face-ID, and anatomy items are forbidden here (they belong to Inter / Count / ID / Anat). Scene-wallpaper questions are excluded from the main Instr aggregate.

\textbf{Stage B (eval; answer only).}
A VLM answers each frozen item on the generation with $\{\mathrm{yes}$/$\mathrm{partial}$/$\mathrm{no}$ and cannot rewrite questions. The public Instr mean aggregates role / asymmetry / prop subtypes with fixed weights (asymm $0.50$, role $0.35$, prop $0.15$). Banks and answering scripts ship with the release.

\paragraph{Example bank (one sample).}
Table~\ref{tab:supp_instr_ex} shows three main-table items from the frozen bank of sample \texttt{hug\_\_69628680878d\_\_T2}. Questions were authored from the edit text alone; Stage~B only answers them on the generation. They are not \texttt{vlm\_judge} checklist fields.

\begin{table}[ht]
\centering
\small
\setlength{\tabcolsep}{4pt}
\begin{tabular}{@{}lp{0.62\textwidth}c@{}}
\toprule
Subtype & Question & Gold \\
\midrule
\texttt{asymm}
  & Is R2 resting his cheek against R1's cheek during the close embrace,
    rather than R1 resting hers against R2's?
  & yes \\
\texttt{role\_duty}
  & Is R1 in tight affectionate body contact with R2 while R2's cheek
    is pressed or rested against hers?
  & yes \\
\texttt{prop\_object}
  & Does R1 hold a yellow plaid gift box?
  & yes \\
\bottomrule
\end{tabular}
\caption{Instruction QA example (one frozen bank; main-table subtypes only).}
\label{tab:supp_instr_ex}
\end{table}

This separation is intentional: the shared checklist is the right instrument for Count and for showing VLM saturation on contact geometry (V-Anat/V-Inter), while Instr needs a reproducible, text-authored bank that does not invent questions from the image under test.
\section{Metric anchors: GT reference and controlled corruption}
\label{supp:corruption}

\textbf{GT reference.} A full held-out GT pack under the public Anat/Inter scorer is pending in the release; we therefore treat GT as a qualitative upper-anchor check rather than a hard numeric row in the main-paper results table.

\textbf{Controlled mesh corruption.} The main paper summarizes direction checks on $200$ Multi-HMR reconstructions of GT targets: force penetration, separate far under required intent, drop person, and duplicate person all lower Inter under the public algebra. Absolute $\Delta$ values are protocol-dependent; we report monotonicity rather than a single frozen delta recipe. A CPU synthetic two-person mesh suite (same Inter algebra; no HMR) additionally passes $6/6$ required direction checks (penetration, separation, drop-person, forbidden-near, bone/self-collision).

\section{Scene-clustered uncertainty for Anat/Inter}
\label{supp:scene_ci}

Full-set Anat/Inter \emph{point estimates} are those in main-paper Table~\ref{tab:main} (frozen also as \texttt{results/tab\_main.json} in the code/data supplement).
Means nest in $405$ scenes. A paired scene-level bootstrap of Seedream$-$Gemini Inter on shared scenes ($B{=}2000$; resample scenes with replacement) yields $\Delta{=}{+}0.042$ with $95\%$ CI $[0.019,0.064]$ (excludes zero): Seedream's Inter lead is significant under this clustering.
Gaps of ${\approx}0.01$ in Table~\ref{tab:main} should be read as ties under the same clustering.
We do \emph{not} reprint a per-model CI table here: older bootstrap dumps on incomplete per-model coverage produced means that disagree with Table~\ref{tab:main}, so those intervals are not authoritative for the public freeze.

\section{Sensitivity analyses}
\label{supp:sens}

Offline recomposition of stored Multi-HMR intermediates (no reconstructor re-run) defines the public Anat/Inter scores. Table~\ref{tab:supp_anat_ablate} reports ten-model Anatomy rank Spearman under weight, attached-signature, and attached-threshold ($\ell$) variants relative to the public mix ($0.40/0.20/0.25/0.15$); all variants keep $\rho\ge0.98$, and FireRed / Gemini / Seedream remain the Anat top-3 except under equal weights (BAGEL enters top-3). The AC-style $0.30/0.30/0.20/0.20$ mix keeps $\rho{=}0.988$ with the same top-3; varying $\ell\in\{0.50,0.62,0.75\}$ keeps $\rho{=}1.000$. ROI padding ($0.22\max(H,W)$) enters only Anat leftover/ROI terms---Interaction is pad-invariant by construction. Re-rasterizing leftover terms from stored Multi-HMR geometry on FireRed ($N{=}150$) at pad fractions $\{0.15,0.22,0.30\}$ leaves Inter unchanged and moves Anat by ${+}0.020$ / $0$ / ${-}0.006$ vs.\ the default $0.22$ setting (Table~\ref{tab:supp_pad}), so the public pad sits in a mild, one-sided neighborhood. Inter top-2 (Seedream/Gemini under the current ten-editor table) remains stable under moderate cutoff and intent-mix perturbations. Full numeric ablation dumps ship with the release scripts.

\begin{table}[ht]
\centering
\small
\begin{tabular}{@{}lccc@{}}
\toprule
Pad frac.\ of $\max(H,W)$ & Anat & Inter & $\Delta$Anat vs $0.22$ \\
\midrule
$0.15$ & $0.695$ & $0.733$ & ${+}0.020$ \\
$0.22$ (public) & $0.676$ & $0.733$ & $0$ \\
$0.30$ & $0.670$ & $0.733$ & ${-}0.006$ \\
\bottomrule
\end{tabular}
\caption{ROI pad sensitivity on FireRed ($N{=}150$ subset; public Anat/Inter recipe). Inter is pad-invariant; Anat moves mildly.}
\label{tab:supp_pad}
\end{table}

\begin{table}[ht]
\centering
\small
\begin{tabular}{@{}llc@{}}
\toprule
Variant & Rank Spearman vs.\ base & Anat top-3 \\
\midrule
base ($0.40/0.20/0.25/0.15$) & $1.000$ & FireRed, Gemini, Seedream \\
equal weights ($0.25^{4}$) & $0.976$ & FireRed, Gemini, BAGEL \\
$0.30/0.30/0.20/0.20$ & $0.988$ & FireRed, Gemini, Seedream \\
attach-heavy & $1.000$ & FireRed, Gemini, Seedream \\
drop $s_{\mathrm{struct}}$ (renorm) & $1.000$ & FireRed, Gemini, Seedream \\
attached-signature off & $0.988$ & FireRed, Gemini, Seedream \\
attached-signature stricter & $1.000$ & FireRed, Gemini, Seedream \\
attached $\ell\ge0.50$ & $1.000$ & FireRed, Gemini, Seedream \\
attached $\ell\ge0.75$ & $1.000$ & FireRed, Gemini, Seedream \\
\bottomrule
\end{tabular}
\caption{Offline Anat weight / attached-signature / $\ell$-threshold ablations on the full set (ten editors; anat\_v4\_exp recompose).}
\label{tab:supp_anat_ablate}
\end{table}

\section{Per-category Anat/Inter}
\label{supp:category}

Table~\ref{tab:supp_category} pools mesh Anat/Inter over ten editors by interaction category (studio source tags CHI3D/Harmony4D excluded). Lowest Anatomy means appear on handshake / hug / high-five; lowest Interaction on face-to-face talk / high-five / hug. Talk Inter is partly depressed by non-contact and unspecified-intent prompts; hug/high-five remain hard contact cases where editors still fail geometry.

\paragraph{Locus-critical slice.}
We mark handshake / hand-hold / high-five / arm-around as \emph{locus-critical}: prompts name a part-specific contact that whole-body $s_{\mathrm{prox}}$ cannot verify.
These four categories cover $491/2{,}500$ samples (${\approx}19.6\%$).
Pooling Table~\ref{tab:supp_category} rows gives Inter ${\approx}0.59$ on the locus-critical slice vs.\ ${\approx}0.56$ on the remainder---comparable or slightly higher---so the public Interaction axis does not down-rank part-specific prompts by itself.
As a stricter probe on a fixed $N{=}150$ subset, we replace whole-body $d_s$ by the minimum 3D wrist/hand distance between the two kept humans on locus-critical IDs only (same Inter algebra otherwise). Mean Inter drops by ${\approx}0.005$--$0.011$ over all scored IDs and by ${\approx}0.02$--$0.04$ on the IDs that enter the hand--hand branch (${\approx}27$--$43/150$ per editor with geometry dumps), confirming that body proximity is mildly more lenient than a part-specific hand distance; we keep whole-body $s_{\mathrm{prox}}$ in the public recipe and treat hand--hand as a Supplementary sensitivity.
The gaming proxy in Section~\ref{supp:gaming} (${\approx}1.3\%$) remains the diagnostic for high-proximity / wrong-region residuals.

\paragraph{$n_{\mathrm{exp}}\ge3$.}
Expected person count is $\ge3$ on $399/2{,}500$ samples ($16\%$; max $9$).
Scoring keeps $k{=}n_{\mathrm{exp}}$ and aggregates Interaction by worst kept pair; under-detection (kept $<n_{\mathrm{exp}}$) on this slice is rare in frozen dumps (${\approx}1\%$) and triggers the public halve rule.

\begin{table}[ht]
\centering
\small
\setlength{\tabcolsep}{3.5pt}
\begin{tabular}{@{}lccc@{}}
\toprule
Category & $n$ & Anat & Inter \\
\midrule
face-to-face talk & 1004 & 0.58 & 0.43 \\
handshake & 994 & 0.54 & 0.57 \\
hand-hold & 1480 & 0.62 & 0.72 \\
high-five & 1031 & 0.56 & 0.48 \\
arm-around & 1268 & 0.58 & 0.54 \\
hug & 1982 & 0.56 & 0.48 \\
piggyback & 1389 & 0.58 & 0.49 \\
carry/lift & 1670 & 0.58 & 0.56 \\
dance & 2073 & 0.58 & 0.59 \\
dance-lift & 1276 & 0.63 & 0.69 \\
combat & 1829 & 0.58 & 0.63 \\
grapple & 1376 & 0.61 & 0.60 \\
other & 2642 & 0.62 & 0.53 \\
\bottomrule
\end{tabular}
\caption{Pooled mesh Anat/Inter by interaction category (ten editors; offline anat\_v4\_exp + inter\_v3.1).}
\label{tab:supp_category}
\end{table}

\section{Interaction gaming proxy}
\label{supp:gaming}

Continuous Interaction does not enforce the semantically required contact region. As a diagnostic, we flag required-intent, recon-successful units with high mesh $s_{\mathrm{pen}}$ and $s_{\mathrm{prox}}$ ($\ge0.85$) whose frozen VLM judge marks \texttt{contact\_points}${}={}$no. This proxy fires on ${\approx}1.3\%$ of required-ok units (higher on several open editors, ${\approx}2$--$2.5\%$; ${\approx}0.5\%$ on closed APIs). Removing these units does not change Inter top-2. A dedicated region-condition term beyond $s_{\mathrm{qual}}$ remains future work; the proxy is asymmetric (VLM-saturated contact tags) and underestimates a residual locus failure mode left for future work.

\section{Detection-confidence proxy}
\label{supp:unc}

Public Anat/Inter do not downweight unstable reconstructions. Splitting full-set successes by mean kept-body Multi-HMR detection score at $0.45$ yields lower means in the low-det bin (Anat/Inter ${\approx}0.55/0.46$) than the high-det bin (${\approx}0.65/0.76$), with comparable Anat stds. We report this only as a ranking diagnostic: detection score is not a calibrated mesh uncertainty, and identity-aware person selection is not yet applied.

\section{Multi-seed generation variance}
\label{supp:multiseed}

Primary tables report a single generation per sample. To quantify sampling noise, we regenerate a fixed 150-sample subset for three open-source editors (FLUX.1~Kontext~dev, ACE++, OmniGen2) at seeds $\{0,1,2\}$ and keep only samples with valid judgments under \emph{all} seeds ($n{=}150$ per model). Seed-level Anat/Inter stds of the three seed means remain small ($O(10^{-2})$), well below the closed/open and density gaps in the main tables; per-image noise is larger, as expected for stochastic editors. Absolute seed-level means shift with the Anat/Inter protocol and are therefore omitted here.

\section{Letterbox / normalize probe}
\label{supp:letterbox}

The main paper defines two official mesh tracks: Track~A (vendor-native public six-axis table) and Track~B (letterbox$\to$$1024^{2}$ on the full $2{,}500$-sample set; Table~\ref{tab:supp_letterbox}). Absolute Anat/Inter rise for all ten editors (Anat $\Delta{\approx}{+}0.06$--$0.14$), while open-editor Anat rank Spearman vs.\ Track~A stays $0.89$ (all-ten $0.83$) and the Anat top-3 is unchanged. Track~A remains the deployment-facing public leaderboard; Track~B is the fair-resolution view.

\begin{table}[ht]
\centering
\small
\setlength{\tabcolsep}{3.5pt}
\begin{tabular}{@{}lcc@{}}
\toprule
Model & Anat native / lb ($\Delta$) & Inter native / lb ($\Delta$) \\
\midrule
GPT-Image-2 & $0.566/0.681$ ($+0.114$) & $0.660/0.754$ ($+0.094$) \\
Gemini-3-Pro-Image & $0.630/0.712$ ($+0.082$) & $0.693/0.791$ ($+0.098$) \\
Seedream-5-Pro & $0.599/0.680$ ($+0.081$) & $0.660/0.803$ ($+0.143$) \\
FireRed-Image-Edit & $0.677/0.736$ ($+0.060$) & $0.712/0.788$ ($+0.076$) \\
BAGEL & $0.600/0.696$ ($+0.097$) & $0.661/0.768$ ($+0.108$) \\
UNO & $0.582/0.691$ ($+0.109$) & $0.655/0.732$ ($+0.077$) \\
FLUX.1-Kontext & $0.538/0.682$ ($+0.144$) & $0.634/0.736$ ($+0.102$) \\
DreamO & $0.539/0.663$ ($+0.124$) & $0.626/0.730$ ($+0.104$) \\
OmniGen2 & $0.542/0.678$ ($+0.136$) & $0.640/0.750$ ($+0.109$) \\
ACE++ & $0.535/0.649$ ($+0.114$) & $0.629/0.710$ ($+0.082$) \\
\midrule
Rank Spearman (Anat) & \multicolumn{2}{c}{closed / open / all: $0.50$ / $0.89$ / $0.83$} \\
\bottomrule
\end{tabular}
\caption{Track~B letterbox$\to$$1024$ on the full $2{,}500$-sample set (same Anat/Inter algebra as Track~A). FireRed Track~A public cells stay $0.63/0.60$.}
\label{tab:supp_letterbox}
\end{table}

\section{Alternate mesh frontend (HMR2)}
\label{supp:frontend}

Public rankings freeze Multi-HMR. For audit we also score the same $N{=}150$ IDs with a detect{+}single-person HMR2 frontend under the \emph{public} Anat/Inter recipe (Table~\ref{tab:supp_frontend}). A single alternate frontend is \emph{not} interchangeable with Multi-HMR on editor order; the useful companion is a two-frontend \emph{mean} of editor means, which agrees with Multi-HMR at rank Spearman ${\approx}0.92$ (Anat) / $0.68$ (Inter) and keeps the same Anat top-3 on that slice (Table~\ref{tab:supp_ensemble}). These subset means are \emph{not} Table~\ref{tab:main}---the full-set public leaders remain Gemini (Anat) / Seedream (Inter). We therefore keep Multi-HMR as the primary leaderboard and release the mean-ensemble numbers as an optional multi-frontend view.

\begin{table}[ht]
\centering
\small
\setlength{\tabcolsep}{3.5pt}
\begin{tabular}{@{}lccc@{}}
\toprule
Model & $n$ & $\rho$ Anat & $\rho$ Inter \\
\midrule
ACE++ & 150 & $0.215$ & $0.110$ \\
BAGEL & 150 & $-0.142$ & $0.064$ \\
DreamO & 150 & $-0.100$ & $-0.126$ \\
FireRed-Image-Edit & 150 & $-0.290$ & $-0.150$ \\
FLUX.1-Kontext & 150 & $-0.062$ & $0.084$ \\
Gemini-3-Pro-Image & 150 & $-0.167$ & $-0.145$ \\
GPT-Image-2 & 150 & $-0.067$ & $0.064$ \\
OmniGen2 & 150 & $-0.071$ & $-0.090$ \\
Seedream-5-Pro & 150 & $-0.184$ & $0.034$ \\
UNO & 150 & $-0.216$ & $-0.051$ \\
\midrule
Ten-model rank Spearman & --- & $0.103$ & $0.248$ \\
\bottomrule
\end{tabular}
\caption{HMR2 vs.\ Multi-HMR under the public Anat/Inter recipe ($N{=}150$). Per-row values are sample-wise Spearman; last row is Spearman of the ten editor means.}
\label{tab:supp_frontend}
\end{table}

\begin{table}[ht]
\centering
\small
\setlength{\tabcolsep}{3.5pt}
\begin{tabular}{@{}lcc@{}}
\toprule
Aggregation vs.\ Multi-HMR & $\rho$ Anat & $\rho$ Inter \\
\midrule
HMR2 alone & $0.10$ & $0.25$ \\
Mean ensemble (MHMR{+}HMR2) & $0.92$ & $0.68$ \\
Worst-case $\min$ ensemble & $0.10$ & $0.25$ \\
\bottomrule
\end{tabular}
\caption{Ten-model rank Spearman vs.\ Multi-HMR editor means ($N{=}150$). Mean ensemble is the companion track; $\min$ matches single-frontend HMR2 and is not used publicly.}
\label{tab:supp_ensemble}
\end{table}

\section{Contact-intent parsing}
\label{supp:intent}

On the human-consistency pool ($N{=}170$), keyword/system intent matches human intent on $145/170$ ($85.3\%$). Confusion is dominated by unspecified$\leftrightarrow$required; required$\leftrightarrow$forbidden disagreements are rare ($2$ units). Replacing keyword intent with human intent on the disagreeing units and recomputing $S_{\mathrm{inter}}^{\mathrm{mesh}}$ changes full-set per-model Inter means by ${<}10^{-3}$ and leaves the ten-model Inter ranking unchanged (Spearman $1.00$).

\section{Human-consistency statistics}
\label{supp:ci}

\textbf{Annotators and gold.} Five adult annotators independently score every unit in the $N{=}170$ consistency pool on the shared checklist (I0, I1, Ic, I3, Ir, A1--A5), following written guidelines that forbid collecting personal information beyond anonymous rater IDs. Main-paper gold is the mean over raters on each item (skipping undecidable marks). Per-annotator exports ship with the release.

\textbf{Inter-annotator reliability (Krippendorff's $\alpha$).}
 Table~\ref{tab:supp_iaa} reports per-item Krippendorff's $\alpha$ over the five raters (interval metric on the per-item scores). The unweighted mean across the ten items is $\alpha{=}0.53$. Core contact items are highest (I1~$0.68$, Ic~$0.65$); mid-range items sit near $0.50$--$0.60$; near-ceiling / high-prevalence items (I0, A4) show lower $\alpha$ even when raw percent-agreement is high, which is expected under class imbalance. We therefore treat human gold as a moderately reliable ranking signal rather than a near-perfect oracle, and interpret H--M / H--V correlations in that light.

\begin{table}[ht]
\centering
\small
\begin{tabular}{@{}lcc@{}}
\toprule
Item & Krippendorff's $\alpha$ & Note \\
\midrule
I0 & 0.35 & Near-ceiling; high \%agree, lower $\alpha$ \\
I1 & 0.68 & Core contact (penetration) \\
Ic & 0.65 & Core contact (fit) \\
I3 & 0.58 & Mid--high \\
Ir & 0.55 & Mid--high \\
A1 & 0.60 & Mid--high \\
A2 & 0.50 & Harder; near $0.5$ \\
A3 & 0.55 & Mid \\
A4 & 0.32 & Near-ceiling; same class as I0 \\
A5 & 0.52 & Harder but ${>}0.5$ \\
\midrule
Mean & 0.53 & Unweighted over 10 items \\
\bottomrule
\end{tabular}
\caption{Five-rater Krippendorff's $\alpha$ on the $N{=}170$ consistency pool.}
\label{tab:supp_iaa}
\end{table}

\textbf{Steiger tests.} On the same pool we compare dependent Spearman correlations $\rho(\mathrm{H},\mathrm{M})$ vs.\ $\rho(\mathrm{H},\mathrm{V})$ using Steiger's test with the empirical $\rho(\mathrm{M},\mathrm{V})$. Significant differences ($p{<}0.05$): M$>$V on I3 and Ir; V$>$M on I0. Other M edges are directional but not significant at $0.05$ (A3: $p{\approx}0.09$).

\textbf{Model-level preference anchors.} Averaging $Q_{\mathrm{anat}}/Q_{\mathrm{inter}}$ and mesh Anat/Inter per editor in the consistency pool yields Spearman $\rho(Q_{\mathrm{anat}},S_{\mathrm{anat}}){\approx}0.87$ and $\rho(Q_{\mathrm{inter}},S_{\mathrm{inter}}){\approx}0.80$ across editors (small $n$ of editors; unit-level correlations remain weaker).

Fisher $z$ $95\%$ intervals for main-paper checklist Spearman $\rho$ with $N{=}170$:

\begin{center}
\small
\begin{tabular}{@{}lcccc@{}}
\toprule
Item & $\rho(\mathrm{H},\mathrm{M})$ & CI & $\rho(\mathrm{H},\mathrm{V})$ & CI \\
\midrule
I0 & 0.41 & $[0.28,0.53]$ & 0.77 & $[0.70,0.82]$ \\
I1 & 0.39 & $[0.25,0.51]$ & 0.30 & $[0.16,0.43]$ \\
Ic & 0.54 & $[0.42,0.64]$ & 0.53 & $[0.41,0.63]$ \\
I3 & 0.43 & $[0.30,0.55]$ & 0.09 & $[-0.06,0.24]$ \\
Ir & 0.46 & $[0.33,0.57]$ & 0.17 & $[0.02,0.31]$ \\
A1 & 0.34 & $[0.20,0.47]$ & 0.20 & $[0.05,0.34]$ \\
A2 & 0.24 & $[0.09,0.38]$ & 0.18 & $[0.03,0.32]$ \\
A3 & 0.38 & $[0.24,0.50]$ & 0.21 & $[0.06,0.35]$ \\
A4 & 0.38 & $[0.24,0.50]$ & 0.33 & $[0.19,0.46]$ \\
A5 & 0.26 & $[0.11,0.39]$ & 0.19 & $[0.04,0.33]$ \\
\bottomrule
\end{tabular}
\end{center}

\begin{figure}[ht]
\centering
\includegraphics[width=0.55\linewidth]{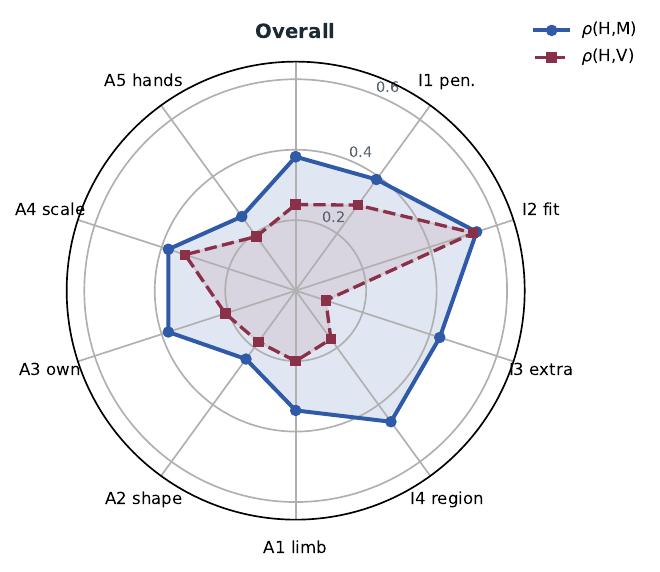}
\caption{Radar view of the per-item $\rho(\mathrm{H},\mathrm{M})$ vs.\ $\rho(\mathrm{H},\mathrm{V})$ table above (main paper Table~\ref{tab:consist_items}).}
\label{fig:supp_consistency}
\end{figure}

\section{Face-Visible Rate and ID conditioning}
\label{supp:fvr}

Mean Face-Visible Rate from \texttt{arcface\_v1} (full scored set):

\begin{center}
\small
\begin{tabular}{@{}lc@{}}
\toprule
Model & FVR \\
\midrule
GPT-Image-2 & 0.91 \\
Gemini-3-Pro-Image & 0.91 \\
Seedream-5-Pro & 0.88 \\
FLUX.1-Kontext & 0.34 \\
DreamO & 0.46 \\
OmniGen2 & 0.39 \\
UNO & 0.15 \\
ACE++ & 0.05 \\
\bottomrule
\end{tabular}
\end{center}

\textbf{ID conditioned on FVR${=}1$.}
To test whether the closed/open identity gap is only hide-the-face, we recompute mean $S_{\mathrm{id}}$ on samples where every GT-visible reference is also matched in the generation ($\mathrm{FVR}\ge0.9999$). Table~\ref{tab:supp_fvr_id} shows a moderate lift ($\approx{+}0.05$--$0.06$), but closed-source ID remains in the mid-$0.5$s--low-$0.6$s. Face visibility explains part of the gap; residual identity failures remain.

\begin{table}[ht]
\centering
\footnotesize
\setlength{\tabcolsep}{3.5pt}
\begin{tabular}{@{}lcccc@{}}
\toprule
Model & $n$ & $n_{{=}1}$ & ID (all) & ID (${=}1$) \\
\midrule
GPT-Image-2 & 1944 & 1645 & 0.576 & 0.638 \\
Gemini-3-Pro-Image & 2321 & 1976 & 0.502 & 0.554 \\
Seedream-5-Pro & 2336 & 1864 & 0.491 & 0.563 \\
\bottomrule
\end{tabular}
\caption{Closed-source identity scores on the full ArcFace-scored set vs.\ the $\mathrm{FVR}{=}1$ subset ($n_{{=}1}$ / ID (${=}1$)).}
\label{tab:supp_fvr_id}
\end{table}

\section{Extended gap list (benchmark survey)}
\label{supp:gap}

Representative suites surveyed when claiming the contact-time anatomy gap (non-exhaustive of every paper in each family; axes marked for Count / ID / Instr / Anat-under-contact / Inter-geometry):

\begin{table}[ht]
\centering
\small
\setlength{\tabcolsep}{3.5pt}
\begin{tabular}{@{}lccccc@{}}
\toprule
Benchmark / suite & Count & ID & Instr & Anat$_c$ & Inter$_g$ \\
\midrule
MultiHuman-Testbench & Y & Y & N & N & N \\
WithAnyone & Y & Y & N & N & N \\
Composing People Together & N & N & N & N & N \\
TrioPose & Y & N & N & N & N \\
GroupDiff & N & Y & N & N & N \\
InsHuman & N & Y & Y & N & N \\
SkeleGuide & N & N & Y & N & N \\
HumanRefiner / ABHUMAN & N & N & N & N$^\dagger$ & N \\
HAF-Bench & N & N & N & N$^\dagger$ & N \\
ImgEdit & N & N & Y & N & N \\
GEdit-Bench & N & N & Y & N & N \\
DreamBench / subject-driven & N & Y & Y & N & N \\
Custom Diff.\ / Dreambooth evals & N & Y & Y & N & N \\
ConceptSlide / multi-concept & N & Y & Y & N & N \\
MagicBrush / EditBench & N & N & Y & N & N \\
TIFA / VQAScore editing probes & N & N & Y & N & N \\
Hi4D / CHI3D (recon.\ GT) & --- & --- & --- & --- & --- \\
BUDDI / proxemics (recon) & --- & --- & --- & --- & --- \\
MPIE-Bench (ours) & Y & Y & Y & Y & Y \\
\bottomrule
\end{tabular}
\caption{Extended benchmark gap survey (Y/N; Anat$_c$~=~anatomy under contact; Inter$_g$~=~interaction geometry).}
\label{tab:supp_gap}
\end{table}
$^\dagger$Single-person anatomy only (no multi-person contact-time scoring).
``---'' denotes reconstruction / GT resources rather than generative editing benchmarks.
The main-paper gap table is the compact name-only view; this extended list covers the broader survey of multi-person, anatomy, personalization, and editing suites.

\section{Release diagnostics}
\label{supp:release}

Per-sample Multi-HMR JSON includes kept detection scores, \texttt{pen\_volume\_m3}, \texttt{pen\_inside\_ratio}, \texttt{min\_surf\_dist}, leftover fractions, and contact intent for offline audit and recomposition.

\end{document}